\documentclass[letterpaper,twocolumn,10pt]{article}
\usepackage{usenix2019_v3}

\usepackage{microtype}
\usepackage{graphicx}
\usepackage{subcaption}
\usepackage{booktabs} %

\usepackage[title]{appendix}
\usepackage{multirow}
\usepackage{hyperref}

\hypersetup{
  colorlinks=true,      %
  linkcolor=black,       %
  citecolor=black,    %
  filecolor=black,       %
  urlcolor=black          %
}

\usepackage{tikz}
\usetikzlibrary{spy}
\usetikzlibrary{patterns}
\usepackage{pgfplots}
\definecolor{color_black_2_0}{rgb}{0.9411764705882353,0.9411764705882353,0.9411764705882353}
\definecolor{color_black_2_1}{rgb}{0.38823529411764707,0.38823529411764707,0.38823529411764707}
\definecolor{color_black_3_0}{rgb}{0.9411764705882353,0.9411764705882353,0.9411764705882353}
\definecolor{color_black_3_1}{rgb}{0.7411764705882353,0.7411764705882353,0.7411764705882353}
\definecolor{color_black_3_2}{rgb}{0.38823529411764707,0.38823529411764707,0.38823529411764707}
\definecolor{color_black_4_0}{rgb}{0.9686274509803922,0.9686274509803922,0.9686274509803922}
\definecolor{color_black_4_1}{rgb}{0.8,0.8,0.8}
\definecolor{color_black_4_2}{rgb}{0.5882352941176471,0.5882352941176471,0.5882352941176471}
\definecolor{color_black_4_3}{rgb}{0.3215686274509804,0.3215686274509804,0.3215686274509804}
\definecolor{color_black_5_0}{rgb}{0.9686274509803922,0.9686274509803922,0.9686274509803922}
\definecolor{color_black_5_1}{rgb}{0.8,0.8,0.8}
\definecolor{color_black_5_2}{rgb}{0.5882352941176471,0.5882352941176471,0.5882352941176471}
\definecolor{color_black_5_3}{rgb}{0.38823529411764707,0.38823529411764707,0.38823529411764707}
\definecolor{color_black_5_4}{rgb}{0.1450980392156863,0.1450980392156863,0.1450980392156863}
\definecolor{color_black_6_0}{rgb}{0.9686274509803922,0.9686274509803922,0.9686274509803922}
\definecolor{color_black_6_1}{rgb}{0.8509803921568627,0.8509803921568627,0.8509803921568627}
\definecolor{color_black_6_2}{rgb}{0.7411764705882353,0.7411764705882353,0.7411764705882353}
\definecolor{color_black_6_3}{rgb}{0.5882352941176471,0.5882352941176471,0.5882352941176471}
\definecolor{color_black_6_4}{rgb}{0.38823529411764707,0.38823529411764707,0.38823529411764707}
\definecolor{color_black_6_5}{rgb}{0.1450980392156863,0.1450980392156863,0.1450980392156863}
\definecolor{color_black_7_0}{rgb}{0.9686274509803922,0.9686274509803922,0.9686274509803922}
\definecolor{color_black_7_1}{rgb}{0.8509803921568627,0.8509803921568627,0.8509803921568627}
\definecolor{color_black_7_2}{rgb}{0.7411764705882353,0.7411764705882353,0.7411764705882353}
\definecolor{color_black_7_3}{rgb}{0.5882352941176471,0.5882352941176471,0.5882352941176471}
\definecolor{color_black_7_4}{rgb}{0.45098039215686275,0.45098039215686275,0.45098039215686275}
\definecolor{color_black_7_5}{rgb}{0.3215686274509804,0.3215686274509804,0.3215686274509804}
\definecolor{color_black_7_6}{rgb}{0.1450980392156863,0.1450980392156863,0.1450980392156863}
\definecolor{color_black_8_0}{rgb}{1.0,1.0,1.0}
\definecolor{color_black_8_1}{rgb}{0.9411764705882353,0.9411764705882353,0.9411764705882353}
\definecolor{color_black_8_2}{rgb}{0.8509803921568627,0.8509803921568627,0.8509803921568627}
\definecolor{color_black_8_3}{rgb}{0.7411764705882353,0.7411764705882353,0.7411764705882353}
\definecolor{color_black_8_4}{rgb}{0.5882352941176471,0.5882352941176471,0.5882352941176471}
\definecolor{color_black_8_5}{rgb}{0.45098039215686275,0.45098039215686275,0.45098039215686275}
\definecolor{color_black_8_6}{rgb}{0.3215686274509804,0.3215686274509804,0.3215686274509804}
\definecolor{color_black_8_7}{rgb}{0.1450980392156863,0.1450980392156863,0.1450980392156863}
\definecolor{color_black_9_0}{rgb}{1.0,1.0,1.0}
\definecolor{color_black_9_1}{rgb}{0.9411764705882353,0.9411764705882353,0.9411764705882353}
\definecolor{color_black_9_2}{rgb}{0.8509803921568627,0.8509803921568627,0.8509803921568627}
\definecolor{color_black_9_3}{rgb}{0.7411764705882353,0.7411764705882353,0.7411764705882353}
\definecolor{color_black_9_4}{rgb}{0.5882352941176471,0.5882352941176471,0.5882352941176471}
\definecolor{color_black_9_5}{rgb}{0.45098039215686275,0.45098039215686275,0.45098039215686275}
\definecolor{color_black_9_6}{rgb}{0.3215686274509804,0.3215686274509804,0.3215686274509804}
\definecolor{color_black_9_7}{rgb}{0.1450980392156863,0.1450980392156863,0.1450980392156863}
\definecolor{color_black_9_8}{rgb}{0.0,0.0,0.0}
\definecolor{color_blue_2_0}{rgb}{0.8705882352941177,0.9215686274509803,0.9686274509803922}
\definecolor{color_blue_2_1}{rgb}{0.19215686274509805,0.5098039215686274,0.7411764705882353}
\definecolor{color_blue_3_0}{rgb}{0.8705882352941177,0.9215686274509803,0.9686274509803922}
\definecolor{color_blue_3_1}{rgb}{0.6196078431372549,0.792156862745098,0.8823529411764706}
\definecolor{color_blue_3_2}{rgb}{0.19215686274509805,0.5098039215686274,0.7411764705882353}
\definecolor{color_blue_4_0}{rgb}{0.9372549019607843,0.9529411764705882,1.0}
\definecolor{color_blue_4_1}{rgb}{0.7411764705882353,0.8431372549019608,0.9058823529411765}
\definecolor{color_blue_4_2}{rgb}{0.4196078431372549,0.6823529411764706,0.8392156862745098}
\definecolor{color_blue_4_3}{rgb}{0.12941176470588237,0.44313725490196076,0.7098039215686275}
\definecolor{color_blue_5_0}{rgb}{0.9372549019607843,0.9529411764705882,1.0}
\definecolor{color_blue_5_1}{rgb}{0.7411764705882353,0.8431372549019608,0.9058823529411765}
\definecolor{color_blue_5_2}{rgb}{0.4196078431372549,0.6823529411764706,0.8392156862745098}
\definecolor{color_blue_5_3}{rgb}{0.19215686274509805,0.5098039215686274,0.7411764705882353}
\definecolor{color_blue_5_4}{rgb}{0.03137254901960784,0.3176470588235294,0.611764705882353}
\definecolor{color_blue_6_0}{rgb}{0.9372549019607843,0.9529411764705882,1.0}
\definecolor{color_blue_6_1}{rgb}{0.7764705882352941,0.8588235294117647,0.9372549019607843}
\definecolor{color_blue_6_2}{rgb}{0.6196078431372549,0.792156862745098,0.8823529411764706}
\definecolor{color_blue_6_3}{rgb}{0.4196078431372549,0.6823529411764706,0.8392156862745098}
\definecolor{color_blue_6_4}{rgb}{0.19215686274509805,0.5098039215686274,0.7411764705882353}
\definecolor{color_blue_6_5}{rgb}{0.03137254901960784,0.3176470588235294,0.611764705882353}
\definecolor{color_blue_7_0}{rgb}{0.9372549019607843,0.9529411764705882,1.0}
\definecolor{color_blue_7_1}{rgb}{0.7764705882352941,0.8588235294117647,0.9372549019607843}
\definecolor{color_blue_7_2}{rgb}{0.6196078431372549,0.792156862745098,0.8823529411764706}
\definecolor{color_blue_7_3}{rgb}{0.4196078431372549,0.6823529411764706,0.8392156862745098}
\definecolor{color_blue_7_4}{rgb}{0.25882352941176473,0.5725490196078431,0.7764705882352941}
\definecolor{color_blue_7_5}{rgb}{0.12941176470588237,0.44313725490196076,0.7098039215686275}
\definecolor{color_blue_7_6}{rgb}{0.03137254901960784,0.27058823529411763,0.5803921568627451}
\definecolor{color_blue_8_0}{rgb}{0.9686274509803922,0.984313725490196,1.0}
\definecolor{color_blue_8_1}{rgb}{0.8705882352941177,0.9215686274509803,0.9686274509803922}
\definecolor{color_blue_8_2}{rgb}{0.7764705882352941,0.8588235294117647,0.9372549019607843}
\definecolor{color_blue_8_3}{rgb}{0.6196078431372549,0.792156862745098,0.8823529411764706}
\definecolor{color_blue_8_4}{rgb}{0.4196078431372549,0.6823529411764706,0.8392156862745098}
\definecolor{color_blue_8_5}{rgb}{0.25882352941176473,0.5725490196078431,0.7764705882352941}
\definecolor{color_blue_8_6}{rgb}{0.12941176470588237,0.44313725490196076,0.7098039215686275}
\definecolor{color_blue_8_7}{rgb}{0.03137254901960784,0.27058823529411763,0.5803921568627451}
\definecolor{color_blue_9_0}{rgb}{0.9686274509803922,0.984313725490196,1.0}
\definecolor{color_blue_9_1}{rgb}{0.8705882352941177,0.9215686274509803,0.9686274509803922}
\definecolor{color_blue_9_2}{rgb}{0.7764705882352941,0.8588235294117647,0.9372549019607843}
\definecolor{color_blue_9_3}{rgb}{0.6196078431372549,0.792156862745098,0.8823529411764706}
\definecolor{color_blue_9_4}{rgb}{0.4196078431372549,0.6823529411764706,0.8392156862745098}
\definecolor{color_blue_9_5}{rgb}{0.25882352941176473,0.5725490196078431,0.7764705882352941}
\definecolor{color_blue_9_6}{rgb}{0.12941176470588237,0.44313725490196076,0.7098039215686275}
\definecolor{color_blue_9_7}{rgb}{0.03137254901960784,0.3176470588235294,0.611764705882353}
\definecolor{color_blue_9_8}{rgb}{0.03137254901960784,0.18823529411764706,0.4196078431372549}
\definecolor{color_green_2_0}{rgb}{0.8980392156862745,0.9607843137254902,0.8784313725490196}
\definecolor{color_green_2_1}{rgb}{0.19215686274509805,0.6392156862745098,0.32941176470588235}
\definecolor{color_green_3_0}{rgb}{0.8980392156862745,0.9607843137254902,0.8784313725490196}
\definecolor{color_green_3_1}{rgb}{0.6313725490196078,0.8509803921568627,0.6078431372549019}
\definecolor{color_green_3_2}{rgb}{0.19215686274509805,0.6392156862745098,0.32941176470588235}
\definecolor{color_green_4_0}{rgb}{0.9294117647058824,0.9725490196078431,0.9137254901960784}
\definecolor{color_green_4_1}{rgb}{0.7294117647058823,0.8941176470588236,0.7019607843137254}
\definecolor{color_green_4_2}{rgb}{0.4549019607843137,0.7686274509803922,0.4627450980392157}
\definecolor{color_green_4_3}{rgb}{0.13725490196078433,0.5450980392156862,0.27058823529411763}
\definecolor{color_green_5_0}{rgb}{0.9294117647058824,0.9725490196078431,0.9137254901960784}
\definecolor{color_green_5_1}{rgb}{0.7294117647058823,0.8941176470588236,0.7019607843137254}
\definecolor{color_green_5_2}{rgb}{0.4549019607843137,0.7686274509803922,0.4627450980392157}
\definecolor{color_green_5_3}{rgb}{0.19215686274509805,0.6392156862745098,0.32941176470588235}
\definecolor{color_green_5_4}{rgb}{0.0,0.42745098039215684,0.17254901960784313}
\definecolor{color_green_6_0}{rgb}{0.9294117647058824,0.9725490196078431,0.9137254901960784}
\definecolor{color_green_6_1}{rgb}{0.7803921568627451,0.9137254901960784,0.7529411764705882}
\definecolor{color_green_6_2}{rgb}{0.6313725490196078,0.8509803921568627,0.6078431372549019}
\definecolor{color_green_6_3}{rgb}{0.4549019607843137,0.7686274509803922,0.4627450980392157}
\definecolor{color_green_6_4}{rgb}{0.19215686274509805,0.6392156862745098,0.32941176470588235}
\definecolor{color_green_6_5}{rgb}{0.0,0.42745098039215684,0.17254901960784313}
\definecolor{color_green_7_0}{rgb}{0.9294117647058824,0.9725490196078431,0.9137254901960784}
\definecolor{color_green_7_1}{rgb}{0.7803921568627451,0.9137254901960784,0.7529411764705882}
\definecolor{color_green_7_2}{rgb}{0.6313725490196078,0.8509803921568627,0.6078431372549019}
\definecolor{color_green_7_3}{rgb}{0.4549019607843137,0.7686274509803922,0.4627450980392157}
\definecolor{color_green_7_4}{rgb}{0.2549019607843137,0.6705882352941176,0.36470588235294116}
\definecolor{color_green_7_5}{rgb}{0.13725490196078433,0.5450980392156862,0.27058823529411763}
\definecolor{color_green_7_6}{rgb}{0.0,0.35294117647058826,0.19607843137254902}
\definecolor{color_green_8_0}{rgb}{0.9686274509803922,0.9882352941176471,0.9607843137254902}
\definecolor{color_green_8_1}{rgb}{0.8980392156862745,0.9607843137254902,0.8784313725490196}
\definecolor{color_green_8_2}{rgb}{0.7803921568627451,0.9137254901960784,0.7529411764705882}
\definecolor{color_green_8_3}{rgb}{0.6313725490196078,0.8509803921568627,0.6078431372549019}
\definecolor{color_green_8_4}{rgb}{0.4549019607843137,0.7686274509803922,0.4627450980392157}
\definecolor{color_green_8_5}{rgb}{0.2549019607843137,0.6705882352941176,0.36470588235294116}
\definecolor{color_green_8_6}{rgb}{0.13725490196078433,0.5450980392156862,0.27058823529411763}
\definecolor{color_green_8_7}{rgb}{0.0,0.35294117647058826,0.19607843137254902}
\definecolor{color_green_9_0}{rgb}{0.9686274509803922,0.9882352941176471,0.9607843137254902}
\definecolor{color_green_9_1}{rgb}{0.8980392156862745,0.9607843137254902,0.8784313725490196}
\definecolor{color_green_9_2}{rgb}{0.7803921568627451,0.9137254901960784,0.7529411764705882}
\definecolor{color_green_9_3}{rgb}{0.6313725490196078,0.8509803921568627,0.6078431372549019}
\definecolor{color_green_9_4}{rgb}{0.4549019607843137,0.7686274509803922,0.4627450980392157}
\definecolor{color_green_9_5}{rgb}{0.2549019607843137,0.6705882352941176,0.36470588235294116}
\definecolor{color_green_9_6}{rgb}{0.13725490196078433,0.5450980392156862,0.27058823529411763}
\definecolor{color_green_9_7}{rgb}{0.0,0.42745098039215684,0.17254901960784313}
\definecolor{color_green_9_8}{rgb}{0.0,0.26666666666666666,0.10588235294117647}
\definecolor{color_orange_2_0}{rgb}{0.996078431372549,0.9019607843137255,0.807843137254902}
\definecolor{color_orange_2_1}{rgb}{0.9019607843137255,0.3333333333333333,0.050980392156862744}
\definecolor{color_orange_3_0}{rgb}{0.996078431372549,0.9019607843137255,0.807843137254902}
\definecolor{color_orange_3_1}{rgb}{0.9921568627450981,0.6823529411764706,0.4196078431372549}
\definecolor{color_orange_3_2}{rgb}{0.9019607843137255,0.3333333333333333,0.050980392156862744}
\definecolor{color_orange_4_0}{rgb}{0.996078431372549,0.9294117647058824,0.8705882352941177}
\definecolor{color_orange_4_1}{rgb}{0.9921568627450981,0.7450980392156863,0.5215686274509804}
\definecolor{color_orange_4_2}{rgb}{0.9921568627450981,0.5529411764705883,0.23529411764705882}
\definecolor{color_orange_4_3}{rgb}{0.8509803921568627,0.2784313725490196,0.00392156862745098}
\definecolor{color_orange_5_0}{rgb}{0.996078431372549,0.9294117647058824,0.8705882352941177}
\definecolor{color_orange_5_1}{rgb}{0.9921568627450981,0.7450980392156863,0.5215686274509804}
\definecolor{color_orange_5_2}{rgb}{0.9921568627450981,0.5529411764705883,0.23529411764705882}
\definecolor{color_orange_5_3}{rgb}{0.9019607843137255,0.3333333333333333,0.050980392156862744}
\definecolor{color_orange_5_4}{rgb}{0.6509803921568628,0.21176470588235294,0.011764705882352941}
\definecolor{color_orange_6_0}{rgb}{0.996078431372549,0.9294117647058824,0.8705882352941177}
\definecolor{color_orange_6_1}{rgb}{0.9921568627450981,0.8156862745098039,0.6352941176470588}
\definecolor{color_orange_6_2}{rgb}{0.9921568627450981,0.6823529411764706,0.4196078431372549}
\definecolor{color_orange_6_3}{rgb}{0.9921568627450981,0.5529411764705883,0.23529411764705882}
\definecolor{color_orange_6_4}{rgb}{0.9019607843137255,0.3333333333333333,0.050980392156862744}
\definecolor{color_orange_6_5}{rgb}{0.6509803921568628,0.21176470588235294,0.011764705882352941}
\definecolor{color_orange_7_0}{rgb}{0.996078431372549,0.9294117647058824,0.8705882352941177}
\definecolor{color_orange_7_1}{rgb}{0.9921568627450981,0.8156862745098039,0.6352941176470588}
\definecolor{color_orange_7_2}{rgb}{0.9921568627450981,0.6823529411764706,0.4196078431372549}
\definecolor{color_orange_7_3}{rgb}{0.9921568627450981,0.5529411764705883,0.23529411764705882}
\definecolor{color_orange_7_4}{rgb}{0.9450980392156862,0.4117647058823529,0.07450980392156863}
\definecolor{color_orange_7_5}{rgb}{0.8509803921568627,0.2823529411764706,0.00392156862745098}
\definecolor{color_orange_7_6}{rgb}{0.5490196078431373,0.17647058823529413,0.01568627450980392}
\definecolor{color_orange_8_0}{rgb}{1.0,0.9607843137254902,0.9215686274509803}
\definecolor{color_orange_8_1}{rgb}{0.996078431372549,0.9019607843137255,0.807843137254902}
\definecolor{color_orange_8_2}{rgb}{0.9921568627450981,0.8156862745098039,0.6352941176470588}
\definecolor{color_orange_8_3}{rgb}{0.9921568627450981,0.6823529411764706,0.4196078431372549}
\definecolor{color_orange_8_4}{rgb}{0.9921568627450981,0.5529411764705883,0.23529411764705882}
\definecolor{color_orange_8_5}{rgb}{0.9450980392156862,0.4117647058823529,0.07450980392156863}
\definecolor{color_orange_8_6}{rgb}{0.8509803921568627,0.2823529411764706,0.00392156862745098}
\definecolor{color_orange_8_7}{rgb}{0.5490196078431373,0.17647058823529413,0.01568627450980392}
\definecolor{color_orange_9_0}{rgb}{1.0,0.9607843137254902,0.9215686274509803}
\definecolor{color_orange_9_1}{rgb}{0.996078431372549,0.9019607843137255,0.807843137254902}
\definecolor{color_orange_9_2}{rgb}{0.9921568627450981,0.8156862745098039,0.6352941176470588}
\definecolor{color_orange_9_3}{rgb}{0.9921568627450981,0.6823529411764706,0.4196078431372549}
\definecolor{color_orange_9_4}{rgb}{0.9921568627450981,0.5529411764705883,0.23529411764705882}
\definecolor{color_orange_9_5}{rgb}{0.9450980392156862,0.4117647058823529,0.07450980392156863}
\definecolor{color_orange_9_6}{rgb}{0.8509803921568627,0.2823529411764706,0.00392156862745098}
\definecolor{color_orange_9_7}{rgb}{0.6509803921568628,0.21176470588235294,0.011764705882352941}
\definecolor{color_orange_9_8}{rgb}{0.4980392156862745,0.15294117647058825,0.01568627450980392}
\definecolor{color_purple_2_0}{rgb}{0.9372549019607843,0.9294117647058824,0.9607843137254902}
\definecolor{color_purple_2_1}{rgb}{0.4588235294117647,0.4196078431372549,0.6941176470588235}
\definecolor{color_purple_3_0}{rgb}{0.9372549019607843,0.9294117647058824,0.9607843137254902}
\definecolor{color_purple_3_1}{rgb}{0.7372549019607844,0.7411764705882353,0.8627450980392157}
\definecolor{color_purple_3_2}{rgb}{0.4588235294117647,0.4196078431372549,0.6941176470588235}
\definecolor{color_purple_4_0}{rgb}{0.9490196078431372,0.9411764705882353,0.9686274509803922}
\definecolor{color_purple_4_1}{rgb}{0.796078431372549,0.788235294117647,0.8862745098039215}
\definecolor{color_purple_4_2}{rgb}{0.6196078431372549,0.6039215686274509,0.7843137254901961}
\definecolor{color_purple_4_3}{rgb}{0.41568627450980394,0.3176470588235294,0.6392156862745098}
\definecolor{color_purple_5_0}{rgb}{0.9490196078431372,0.9411764705882353,0.9686274509803922}
\definecolor{color_purple_5_1}{rgb}{0.796078431372549,0.788235294117647,0.8862745098039215}
\definecolor{color_purple_5_2}{rgb}{0.6196078431372549,0.6039215686274509,0.7843137254901961}
\definecolor{color_purple_5_3}{rgb}{0.4588235294117647,0.4196078431372549,0.6941176470588235}
\definecolor{color_purple_5_4}{rgb}{0.32941176470588235,0.15294117647058825,0.5607843137254902}
\definecolor{color_purple_6_0}{rgb}{0.9490196078431372,0.9411764705882353,0.9686274509803922}
\definecolor{color_purple_6_1}{rgb}{0.8549019607843137,0.8549019607843137,0.9215686274509803}
\definecolor{color_purple_6_2}{rgb}{0.7372549019607844,0.7411764705882353,0.8627450980392157}
\definecolor{color_purple_6_3}{rgb}{0.6196078431372549,0.6039215686274509,0.7843137254901961}
\definecolor{color_purple_6_4}{rgb}{0.4588235294117647,0.4196078431372549,0.6941176470588235}
\definecolor{color_purple_6_5}{rgb}{0.32941176470588235,0.15294117647058825,0.5607843137254902}
\definecolor{color_purple_7_0}{rgb}{0.9490196078431372,0.9411764705882353,0.9686274509803922}
\definecolor{color_purple_7_1}{rgb}{0.8549019607843137,0.8549019607843137,0.9215686274509803}
\definecolor{color_purple_7_2}{rgb}{0.7372549019607844,0.7411764705882353,0.8627450980392157}
\definecolor{color_purple_7_3}{rgb}{0.6196078431372549,0.6039215686274509,0.7843137254901961}
\definecolor{color_purple_7_4}{rgb}{0.5019607843137255,0.49019607843137253,0.7294117647058823}
\definecolor{color_purple_7_5}{rgb}{0.41568627450980394,0.3176470588235294,0.6392156862745098}
\definecolor{color_purple_7_6}{rgb}{0.2901960784313726,0.0784313725490196,0.5254901960784314}
\definecolor{color_purple_8_0}{rgb}{0.9882352941176471,0.984313725490196,0.9921568627450981}
\definecolor{color_purple_8_1}{rgb}{0.9372549019607843,0.9294117647058824,0.9607843137254902}
\definecolor{color_purple_8_2}{rgb}{0.8549019607843137,0.8549019607843137,0.9215686274509803}
\definecolor{color_purple_8_3}{rgb}{0.7372549019607844,0.7411764705882353,0.8627450980392157}
\definecolor{color_purple_8_4}{rgb}{0.6196078431372549,0.6039215686274509,0.7843137254901961}
\definecolor{color_purple_8_5}{rgb}{0.5019607843137255,0.49019607843137253,0.7294117647058823}
\definecolor{color_purple_8_6}{rgb}{0.41568627450980394,0.3176470588235294,0.6392156862745098}
\definecolor{color_purple_8_7}{rgb}{0.2901960784313726,0.0784313725490196,0.5254901960784314}
\definecolor{color_purple_9_0}{rgb}{0.9882352941176471,0.984313725490196,0.9921568627450981}
\definecolor{color_purple_9_1}{rgb}{0.9372549019607843,0.9294117647058824,0.9607843137254902}
\definecolor{color_purple_9_2}{rgb}{0.8549019607843137,0.8549019607843137,0.9215686274509803}
\definecolor{color_purple_9_3}{rgb}{0.7372549019607844,0.7411764705882353,0.8627450980392157}
\definecolor{color_purple_9_4}{rgb}{0.6196078431372549,0.6039215686274509,0.7843137254901961}
\definecolor{color_purple_9_5}{rgb}{0.5019607843137255,0.49019607843137253,0.7294117647058823}
\definecolor{color_purple_9_6}{rgb}{0.41568627450980394,0.3176470588235294,0.6392156862745098}
\definecolor{color_purple_9_7}{rgb}{0.32941176470588235,0.15294117647058825,0.5607843137254902}
\definecolor{color_purple_9_8}{rgb}{0.24705882352941178,0.0,0.49019607843137253}
\definecolor{color_red_2_0}{rgb}{0.996078431372549,0.8784313725490196,0.8235294117647058}
\definecolor{color_red_2_1}{rgb}{0.8705882352941177,0.17647058823529413,0.14901960784313725}
\definecolor{color_red_3_0}{rgb}{0.996078431372549,0.8784313725490196,0.8235294117647058}
\definecolor{color_red_3_1}{rgb}{0.9882352941176471,0.5725490196078431,0.4470588235294118}
\definecolor{color_red_3_2}{rgb}{0.8705882352941177,0.17647058823529413,0.14901960784313725}
\definecolor{color_red_4_0}{rgb}{0.996078431372549,0.8980392156862745,0.8509803921568627}
\definecolor{color_red_4_1}{rgb}{0.9882352941176471,0.6823529411764706,0.5686274509803921}
\definecolor{color_red_4_2}{rgb}{0.984313725490196,0.41568627450980394,0.2901960784313726}
\definecolor{color_red_4_3}{rgb}{0.796078431372549,0.09411764705882353,0.11372549019607843}
\definecolor{color_red_5_0}{rgb}{0.996078431372549,0.8980392156862745,0.8509803921568627}
\definecolor{color_red_5_1}{rgb}{0.9882352941176471,0.6823529411764706,0.5686274509803921}
\definecolor{color_red_5_2}{rgb}{0.984313725490196,0.41568627450980394,0.2901960784313726}
\definecolor{color_red_5_3}{rgb}{0.8705882352941177,0.17647058823529413,0.14901960784313725}
\definecolor{color_red_5_4}{rgb}{0.6470588235294118,0.058823529411764705,0.08235294117647059}
\definecolor{color_red_6_0}{rgb}{0.996078431372549,0.8980392156862745,0.8509803921568627}
\definecolor{color_red_6_1}{rgb}{0.9882352941176471,0.7333333333333333,0.6313725490196078}
\definecolor{color_red_6_2}{rgb}{0.9882352941176471,0.5725490196078431,0.4470588235294118}
\definecolor{color_red_6_3}{rgb}{0.984313725490196,0.41568627450980394,0.2901960784313726}
\definecolor{color_red_6_4}{rgb}{0.8705882352941177,0.17647058823529413,0.14901960784313725}
\definecolor{color_red_6_5}{rgb}{0.6470588235294118,0.058823529411764705,0.08235294117647059}
\definecolor{color_red_7_0}{rgb}{0.996078431372549,0.8980392156862745,0.8509803921568627}
\definecolor{color_red_7_1}{rgb}{0.9882352941176471,0.7333333333333333,0.6313725490196078}
\definecolor{color_red_7_2}{rgb}{0.9882352941176471,0.5725490196078431,0.4470588235294118}
\definecolor{color_red_7_3}{rgb}{0.984313725490196,0.41568627450980394,0.2901960784313726}
\definecolor{color_red_7_4}{rgb}{0.9372549019607843,0.23137254901960785,0.17254901960784313}
\definecolor{color_red_7_5}{rgb}{0.796078431372549,0.09411764705882353,0.11372549019607843}
\definecolor{color_red_7_6}{rgb}{0.6,0.0,0.050980392156862744}
\definecolor{color_red_8_0}{rgb}{1.0,0.9607843137254902,0.9411764705882353}
\definecolor{color_red_8_1}{rgb}{0.996078431372549,0.8784313725490196,0.8235294117647058}
\definecolor{color_red_8_2}{rgb}{0.9882352941176471,0.7333333333333333,0.6313725490196078}
\definecolor{color_red_8_3}{rgb}{0.9882352941176471,0.5725490196078431,0.4470588235294118}
\definecolor{color_red_8_4}{rgb}{0.984313725490196,0.41568627450980394,0.2901960784313726}
\definecolor{color_red_8_5}{rgb}{0.9372549019607843,0.23137254901960785,0.17254901960784313}
\definecolor{color_red_8_6}{rgb}{0.796078431372549,0.09411764705882353,0.11372549019607843}
\definecolor{color_red_8_7}{rgb}{0.6,0.0,0.050980392156862744}
\definecolor{color_red_9_0}{rgb}{1.0,0.9607843137254902,0.9411764705882353}
\definecolor{color_red_9_1}{rgb}{0.996078431372549,0.8784313725490196,0.8235294117647058}
\definecolor{color_red_9_2}{rgb}{0.9882352941176471,0.7333333333333333,0.6313725490196078}
\definecolor{color_red_9_3}{rgb}{0.9882352941176471,0.5725490196078431,0.4470588235294118}
\definecolor{color_red_9_4}{rgb}{0.984313725490196,0.41568627450980394,0.2901960784313726}
\definecolor{color_red_9_5}{rgb}{0.9372549019607843,0.23137254901960785,0.17254901960784313}
\definecolor{color_red_9_6}{rgb}{0.796078431372549,0.09411764705882353,0.11372549019607843}
\definecolor{color_red_9_7}{rgb}{0.6470588235294118,0.058823529411764705,0.08235294117647059}
\definecolor{color_red_9_8}{rgb}{0.403921568627451,0.0,0.050980392156862744}
\usepackage{xspace}

\newcommand{\paragraphOrTextbf}[1]{\textbf{#1}\xspace}

\newcommand{\ourtitle}{\sys: Efficient Fault Tolerance for Recommendation Model Training\\via Erasure Coding}

\usepackage{amsmath}
\usepackage{nccmath}
\usepackage{xpatch}
\xpatchcmd{\NCC@ignorepar}{%
\abovedisplayskip\abovedisplayshortskip}
{%
\abovedisplayskip\abovedisplayshortskip%
\belowdisplayskip\belowdisplayshortskip}
{}{}

\usepackage{enumitem}
\newenvironment{denseitemize}{
\begin{itemize}[topsep=2.5pt, partopsep=0pt, leftmargin=1.5em]
  \setlength{\itemsep}{2.5pt}
  \setlength{\parskip}{0pt}
  \setlength{\parsep}{0pt}
}{\end{itemize}}

\newcommand{\ie}{i.e.,\xspace}
\newcommand{\eg}{e.g.,\xspace}

\newcommand{\Section}{\S}
\newcommand{\Figure}{Figure~}

\newcommand{\sys}{ECRM\xspace}

\newcommand{\dlrm}{DLRM\xspace}
\newcommand{\dlrms}{DLRMs\xspace}

\newcommand{\nn}{neural network\xspace}
\newcommand{\nns}{neural networks\xspace}

\newcommand{\nnShort}{neural network\xspace}
\newcommand{\nnsShort}{neural networks\xspace}

\newcommand{\xdl}{XDL\xspace}

\newcommand{\embeddingTable}{embedding table\xspace}
\newcommand{\embeddingTables}{embedding tables\xspace}

\newcommand{\EmbeddingTables}{Embedding tables\xspace}

\newcommand{\entry}{entry\xspace}
\newcommand{\entries}{entries\xspace}
\newcommand{\Entry}{Entry\xspace}

\newcommand{\embeddingTableEntry}{\embeddingTable \entry}
\newcommand{\embeddingTableEntries}{\embeddingTable \entries}

\newcommand{\server}{server\xspace}
\newcommand{\servers}{servers\xspace}
\newcommand{\Server}{Server\xspace}
\newcommand{\Servers}{Servers\xspace}
\newcommand{\worker}{worker\xspace}
\newcommand{\workers}{workers\xspace}

\newcommand{\Workers}{Workers\xspace}
\newcommand{\optimizer}{optimizer\xspace}
\newcommand{\optimizers}{optimizers\xspace}
\newcommand{\Optimizer}{Optimizer\xspace}

\newcommand{\update}{update\xspace}
\newcommand{\updates}{updates\xspace}

\newcommand{\accumulator}{accumulator\xspace}

\newcommand{\parityAccumulator}{parity \accumulator}

\newcommand{\parityEntry}{parity \entry}
\newcommand{\parityEntries}{parity \entries}

\newcommand{\expWorkerType}{p3.2xlarge\xspace}

\newcommand{\myTime}{t\xspace}
\newcommand{\timePlusOne}{\myTime + 1\xspace}
\newcommand{\timeZero}{0\xspace}
\newcommand{\timeOne}{1\xspace}

\newcommand{\idxOne}{0\xspace}
\newcommand{\idxTwo}{1\xspace}
\newcommand{\idxThree}{2\xspace}
\newcommand{\idxI}{i\xspace}

\newcommand{\lr}{\alpha\xspace}

\newcommand{\emb}{e\xspace}
\newcommand{\embOne}{\emb_\idxOne\xspace}
\newcommand{\embTwo}{\emb_\idxTwo\xspace}
\newcommand{\embThree}{\emb_\idxThree\xspace}
\newcommand{\embI}{\emb_\idxI\xspace}
\newcommand{\grad}{\nabla\xspace}
\newcommand{\gradOne}{\grad_\idxOne\xspace}
\newcommand{\gradTwo}{\grad_\idxTwo\xspace}

\newcommand{\parity}{p\xspace}

\newcommand{\embOneT}{\emb_{\idxOne,\myTime}\xspace}

\newcommand{\embIT}{\emb_{\idxI,\myTime}\xspace}
\newcommand{\gradOneT}{\grad_{\idxOne,\myTime}\xspace}

\newcommand{\gradIT}{\grad_{\idxI,\myTime}\xspace}

\newcommand{\embOneTPlusOne}{\emb_{\idxOne,{\timePlusOne}}\xspace}

\newcommand{\gradOneTZero}{\grad_{\idxOne,{\timeZero}}\xspace}

\newcommand{\gradOneTOne}{\grad_{\idxOne,{\timeOne}}\xspace}

\newcommand{\adagradAccumOne}{G_{\idxOne}\xspace}
\newcommand{\adagradAccumTwo}{G_{\idxTwo}\xspace}
\newcommand{\adagradAccumThree}{G_{\idxThree}\xspace}
\newcommand{\adagradAccumP}{G_{\parity}\xspace}

\newcommand{\adagradAccumOneT}{G_{\idxOne,\myTime}\xspace}

\newcommand{\adagradAccumPT}{G_{\parity,\myTime}\xspace}

\newcommand{\adagradEps}{\epsilon\xspace}

\newcommand{\serverSideProp}{difference propagation\xspace}

\newcommand{\ServerSideProp}{Difference propagation\xspace}

\newcommand{\workerSideProp}{gradient propagation\xspace}

\newcommand{\criteo}{Criteo\xspace}
\newcommand{\criteoOg}{\criteo-Original\xspace}
\newcommand{\criteoTwoSparse}{\criteo-2S\xspace}
\newcommand{\criteoTwoSparseTwoDense}{\criteo-2S-2D\xspace}
\newcommand{\criteoOgSize}{220\xspace}
\newcommand{\criteoTwoSparseSize}{440\xspace}
\newcommand{\criteoTwoSparseTwoDenseSize}{880\xspace}
\newcommand{\criteoOgSizeGB}{\criteoOgSize GB\xspace}
\newcommand{\criteoTwoSparseSizeGB}{\criteoTwoSparseSize GB\xspace}
\newcommand{\criteoTwoSparseTwoDenseSizeGB}{\criteoTwoSparseTwoDenseSize GB\xspace}

\newcommand{\sysKFour}{\sys ($k=4$)\xspace}
\newcommand{\sysKTwo}{\sys ($k=2$)\xspace}
\newcommand{\ckptThirty}{Ckpt-30\xspace}
\newcommand{\ckptSixty}{Ckpt-60\xspace}
\newcommand{\noFT}{No FT\xspace}

\newcommand{\checkpointPeriod}{T\xspace}

\newcommand{\evalNormalCriteoSizeOgCkptFour}{4.559119596308321}
\newcommand{\evalNormalCriteoSizeOgCkptEight}{9.029272659116543}

\newcommand{\evalNormalCriteoSizeOgCodedServerFour}{2.591569924707244} %
\newcommand{\evalNormalCriteoSizeOgCodedServerTwo}{2.3640186823257494}

\newcommand{\evalNormalCriteoSizeTwoSparseCkptFour}{7.85390705097457}
\newcommand{\evalNormalCriteoSizeTwoSparseCkptEight}{17.156594613870254}

\newcommand{\evalNormalCriteoSizeTwoSparseCodedServerFour}{2.6115569737191606} %
\newcommand{\evalNormalCriteoSizeTwoSparseCodedServerTwo}{2.9852772308999107} %

\newcommand{\evalNormalCriteoSizeTwoSparseTwoDenseCkptFour}{14.858179002922142}
\newcommand{\evalNormalCriteoSizeTwoSparseTwoDenseCkptEight}{33.419301912032125}

\newcommand{\evalNormalCriteoSizeTwoSparseTwoDenseCodedServerFour}{4.210906349828917} %
\newcommand{\evalNormalCriteoSizeTwoSparseTwoDenseCodedServerTwo}{4.014091767756653} 

\newcommand{\evalRecoveryCriteoSizeOgCkptFourBest}{1.328} %
\newcommand{\evalRecoveryCriteoSizeOgCkptEightBest}{1.328}
\newcommand{\evalRecoveryCriteoSizeOgCkptFourAverage}{31.328}
\newcommand{\evalRecoveryCriteoSizeOgCkptEightAverage}{16.328}
\newcommand{\evalRecoveryCriteoSizeOgCkptFourWorst}{61.328}
\newcommand{\evalRecoveryCriteoSizeOgCkptEightWorst}{31.328}
\newcommand{\evalRecoveryCriteoSizeOgCodedServerFour}{4.6166666667} %
\newcommand{\evalRecoveryCriteoSizeOgCodedServerTwo}{3.05} %

\newcommand{\evalRecoveryCriteoSizeTwoSparseCkptFourBest}{2.9156666667} %
\newcommand{\evalRecoveryCriteoSizeTwoSparseCkptEightBest}{2.9156666667}
\newcommand{\evalRecoveryCriteoSizeTwoSparseCkptFourAverage}{32.9156666667}
\newcommand{\evalRecoveryCriteoSizeTwoSparseCkptEightAverage}{17.9156666667}
\newcommand{\evalRecoveryCriteoSizeTwoSparseCkptFourWorst}{62.9156666667}
\newcommand{\evalRecoveryCriteoSizeTwoSparseCkptEightWorst}{32.9156666667}
\newcommand{\evalRecoveryCriteoSizeTwoSparseCodedServerFour}{9.5166666667} %
\newcommand{\evalRecoveryCriteoSizeTwoSparseCodedServerTwo}{5.5333333333} %

\newcommand{\evalRecoveryCriteoSizeTwoSparseTwoDenseCkptFourBest}{5.995} %
\newcommand{\evalRecoveryCriteoSizeTwoSparseTwoDenseCkptEightBest}{5.995}
\newcommand{\evalRecoveryCriteoSizeTwoSparseTwoDenseCkptFourAverage}{35.995}
\newcommand{\evalRecoveryCriteoSizeTwoSparseTwoDenseCkptEightAverage}{20.995}
\newcommand{\evalRecoveryCriteoSizeTwoSparseTwoDenseCkptFourWorst}{65.995}
\newcommand{\evalRecoveryCriteoSizeTwoSparseTwoDenseCkptEightWorst}{35.995}
\newcommand{\evalRecoveryCriteoSizeTwoSparseTwoDenseCodedServerFour}{19.0666666667} %
\newcommand{\evalRecoveryCriteoSizeTwoSparseTwoDenseCodedServerTwo}{10.8166666667} %

\newcommand{\ckptOgWriteMinutes}{2.5}
\newcommand{\ckptTwoSparseWriteMinutes}{4.6667}
\newcommand{\ckptTwoSparseTwoDenseWriteMinutes}{9.16667}
\newcommand{\ckptOgReadMinutes}{1.328}
\newcommand{\ckptTwoSparseReadMinutes}{2.9156666667}
\newcommand{\ckptTwoSparseTwoDenseReadMinutes}{5.995}

\begin{document}

\date{}

\title{\Large \bf \ourtitle}

\makeatletter
\newcommand{\printfnsymbol}[1]{%
  \textsuperscript{ \@fnsymbol{#1}}%
}
\makeatother

\author{
{\rm Kaige Liu}\thanks{Work done while at Carnegie Mellon University} \printfnsymbol{2}\\
Facebook\\
kaigeliu98@gmail.com
\and
{\rm Jack Kosaian}\thanks{Equal contribution}\\
Carnegie Mellon University\\
jkosaian@cs.cmu.edu
\and
{\rm K.\ V.\ Rashmi}\\
Carnegie Mellon University\\
rvinayak@cs.cmu.edu
} %

\maketitle

\begin{abstract}
Deep-learning-based recommendation models (DLRMs) are widely deployed to serve personalized content to users. DLRMs are large in size due to their use of large embedding tables, and are trained by  distributing the model across the memory of tens or hundreds of servers. Server failures are common in such large distributed systems and must be mitigated to enable training to progress. Checkpointing is the primary approach used for fault tolerance in these systems, but incurs significant training-time overhead both during normal operation and when recovering from failures. As these overheads increase with DLRM size, checkpointing is slated to become an even larger overhead for future DLRMs, which are expected to grow in size. This calls for rethinking fault tolerance in DLRM training.

We present ECRM, a DLRM training system that achieves efficient fault tolerance using erasure coding. ECRM chooses which DLRM parameters to encode, correctly and efficiently updates parities, and enables training to proceed without any pauses, while maintaining consistency of the recovered parameters. We implement ECRM atop XDL, an open-source, industrial-scale DLRM training system. Compared to checkpointing, ECRM reduces training-time overhead for large DLRMs by up to 88\%, recovers from failures up to 10.3$\times$ faster, and allows training to proceed during recovery. These results show the promise of erasure coding in imparting efficient fault tolerance to training current and future DLRMs.
\end{abstract}
\section{Introduction} \label{sec:intro}
Deep-learning-based recommendation models (\dlrms) are key tools in serving personalized content to users at Internet scale~\cite{covington2016deep,naumov2019deep}. As the value generated by \dlrms often relies on the ability to reflect recent data, production \dlrms are frequently retrained~\cite{nvidia-merlin-hugectr}. Reducing \dlrm training time is thus critical to maintaining an accurate and up-to-date model. 

\dlrms consist of \embeddingTables and \nns. \EmbeddingTables map sparse categorical features (e.g., properties of a client) to a learned dense representation.
\EmbeddingTables resemble lookup tables in which millions or billions~\cite{eisenman2018bandana,jiang2019xdl} of sparse features each map to a small dense vector representation of tens to hundreds of floats. We refer to a single dense embedding vector corresponding to a single sparse feature as an ``\embeddingTableEntry,'' or ``\entry'' for short. A small \nnShort processes dense vectors resulting from \embeddingTable ``lookups'' to produce a final prediction (\eg whether a client like a video).  %

\begin{figure*}[t]
    \centering
    \begin{subfigure}[t]{0.45\textwidth}
        \centering
        \includegraphics[width=0.9\linewidth]{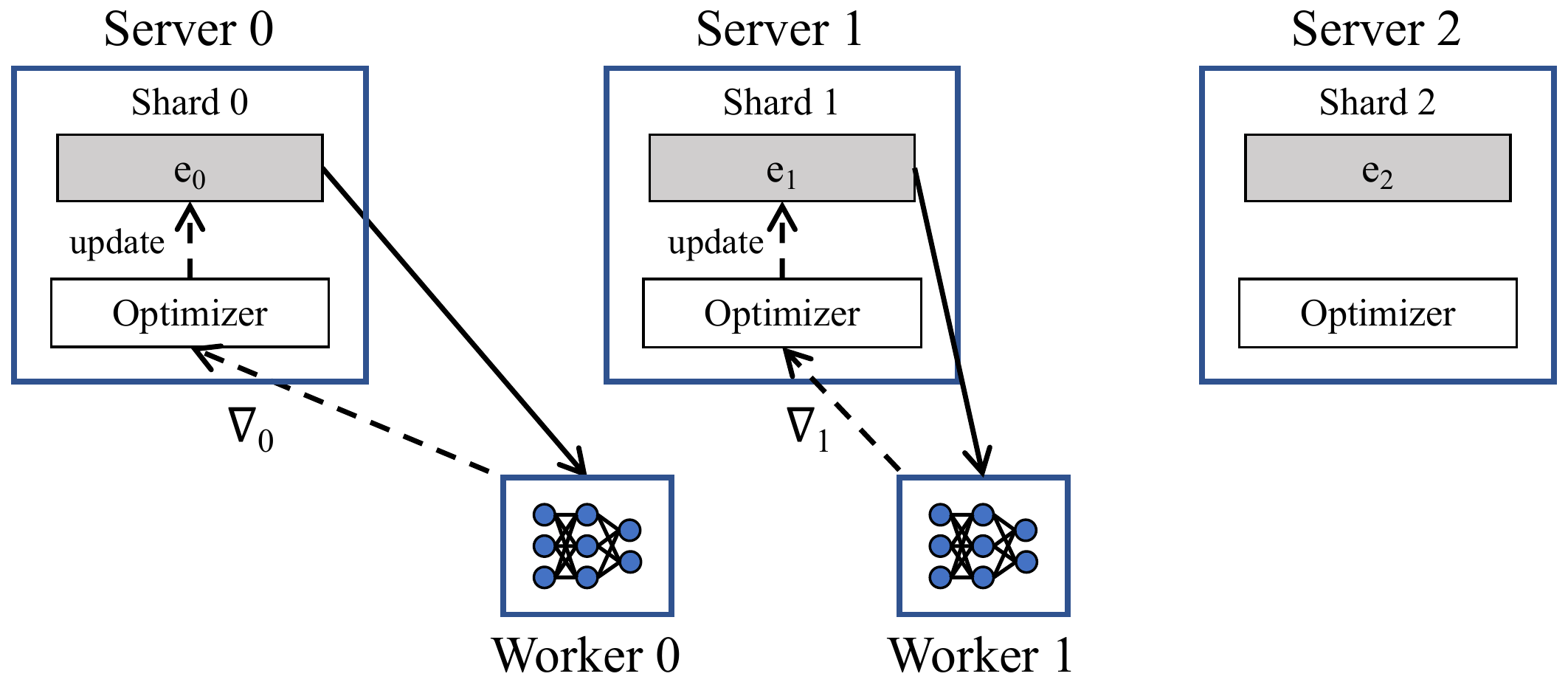}
        \caption{Example of the distributed setup used to train \dlrms.}
        \label{fig:background:embedding}
    \end{subfigure} \hfill
    \begin{subfigure}[t]{0.45\textwidth}
        \centering
        \includegraphics[width=0.9\linewidth]{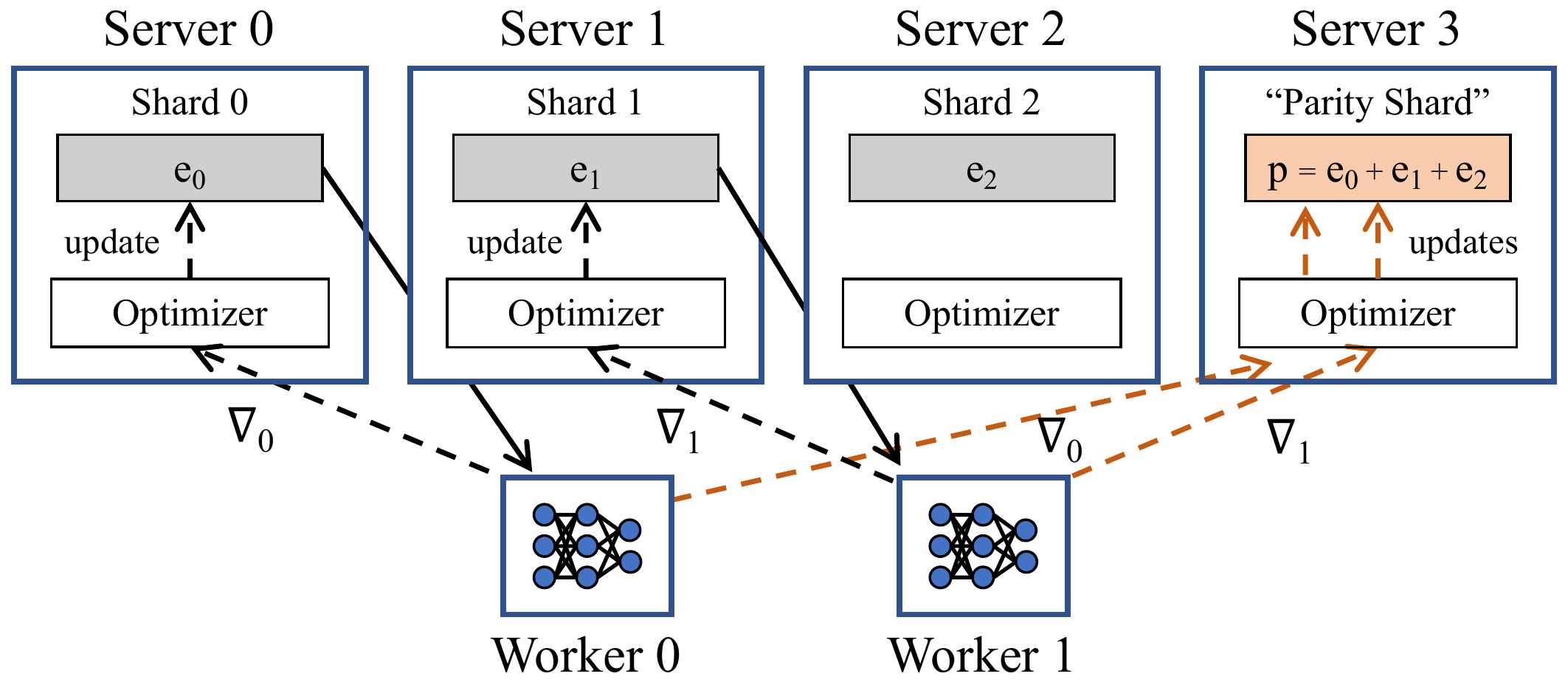}
        \caption{Naive erasure-coded \dlrm with $k=3$ and $r=1$.}
        \label{fig:background:embedding_coded}
        \label{fig:design:update_worker}
    \end{subfigure}
    \vspace{-0.1in}
    \caption{Example of (a) normal and (b) erasure-coded \dlrm training with \embeddingTableEntries $\embOne$, $\embTwo$, and $\embThree$, and gradients $\gradOne$ and $\gradTwo$. 
    }
    \label{fig:background:embedding_compare}
\end{figure*}

\EmbeddingTables are typically large, ranging from hundreds of gigabytes to terabytes in size~\cite{jiang2019xdl}. Such large models are trained in a distributed fashion across tens/hundreds of nodes~\cite{jiang2019xdl,acun2020understanding}, as depicted (at a much smaller scale) in \Figure\ref{fig:background:embedding}. \EmbeddingTables and \nnShort parameters are sharded across a set of \textit{\servers} and kept in memory for fast access. \textit{\Workers} operate in a data-parallel fashion to perform \nnShort training by accessing model parameters from \servers, send gradients to \servers to update parameters via an \optimizer (\eg Adam).

Since model parameters are stored in memory, any server failure requires training to restart from scratch. Given that \dlrm training is resource and time intensive and that failures are common in large-scale settings, it is imperative for \dlrm training to be fault tolerant~\cite{maeng2021cpr}. In this work, we focus on imparting fault tolerance against \server failures. Handling \server failures is critical, as failure of a single \server results in loss of fraction of \embeddingTableEntries. In contrast additional fault tolerance is not needed for handling \worker failures, as each worker contains a replica of the \dlrm's \nn, which can simply be recovered from another worker.

Checkpointing is the main approach used for fault tolerance in \dlrm training~\cite{maeng2021cpr}. This involves periodically pausing training and writing the current parameters and \optimizer state to stable storage, such as a distributed file system. If a failure occurs, the entire system resets to the most recent checkpoint and restarts training from that point. 
While simple, checkpointing frequently pauses training to save \dlrm state and has to redo work after failure. Thus, checkpointing has been shown to add significant overhead to training production \dlrms~\cite{maeng2021cpr}. Even more concerning, this overhead increases with \dlrm size. Given the trend of increasing model size~\cite{huang2020mixed}, checkpointing is slated to incur even larger overhead for future \dlrms.

An alternative to checkpointing that does not require stalls or lengthy recovery is to replicate \dlrm parameters on separate \servers. However, replication requires at least $2\times$ as much memory as a checkpointing-based system, which is impractical given the large sizes of \dlrms. Another alternative is to reduce the overhead of checkpointing by taking approximate checkpoints~\cite{qiao2019fault,chen2020efficient,eisenman2020check,maeng2021cpr}. However, this can result in accuracy loss in training, which makes debugging production systems harder due to uncertainty in the accuracy of the model recovered from the checkpoint. Furthermore, even small drops in accuracy have been noted to result in a significant reduction in the business value generated by \dlrms~\cite{zhao2019aibox,eisenman2020check}, making potential accuracy loss induced by an approach to fault tolerance undesirable.

An ideal approach to fault-tolerant \dlrm training would (1) operate with low training-time overhead, with (2) low memory overhead, while (3) not introducing potential accuracy loss (and the associated uncertainty). Designing such a fault tolerance approach is the goal of this paper.

Erasure codes are coding-theoretic tools for adding proactive redundancy (like replication) but with significantly less memory overhead, which have been widely employed in storage and communication systems (\eg~\cite{patterson1988raid,rizzo1997effective}). Like replication and traditional checkpointing, erasure coding would not alter the accuracy of training. Due to their low memory overhead, erasure codes offer potential for efficient fault tolerance in \dlrm training. As shown in \Figure\ref{fig:background:embedding_coded}, an erasure-coded \dlrm training system would construct ``parity parameters'' by encoding $k$ \embeddingTableEntries from separate \servers. In this example, a parity $\parity$ is formed from parameters $\embOne$, $\embTwo$, and $\embThree$ via the encoding function $\parity = \embOne + \embTwo + \embThree$, and placed on a separate \server. If a \server fails, lost parameters are recovered by reading the $k$ available parameters and performing the erasure code's decoding process (\eg $\embTwo = \parity - \embOne - \embThree$). 

While erasure codes appear promising for imparting fault tolerance to \dlrm training, this vision presents a number of challenges: (1) Parities must be kept up-to-date with \dlrm parameters to ensure correct recovery. This requires additional communication and computation, which can reduce the throughput of training. (2) As will be shown in \Section\ref{sec:challenge:optimizer}, correctly updating parities when using \optimizers that store internal state (\eg Adagrad, Adam) is challenging without incurring large memory overhead. (3) An erasure code's recovery process can be resource intensive~\cite{sathiamoorthy2013xoring,rashmi2014hitchhiker}. This can potentially lead to long recovery times during which training is stalled. 

We present \sys,\footnote{\sys: \underline{E}rasure-\underline{C}oded \underline{R}ecommendation \underline{M}odel} an erasure-coded \dlrm training system that overcomes these challenges through a careful system design adapting simple erasure codes and ideas from storage systems to \dlrm training. 
\sys reduces training-time overhead and circumvents the difficulty of maintaining correctness with stateful \optimizers (Challenges~1 and 2) by delegating the responsibility for updating parities to \servers, rather than \workers, via an approach we call ``\serverSideProp.'' \sys recovers quickly from failure (Challenge~3) by enabling training to continue during the erasure code's recovery process. The net result of \sys's design is a \dlrm training system that \textit{recovers quickly from failures with low training-time and memory overhead, without requiring pauses during normal operation or recovery, and without any changing the accuracy guarantees of the underlying training system.}

We implement \sys atop \xdl, an open-source, industrial-scale \dlrm training system developed by Alibaba~\cite{jiang2019xdl}. We evaluate \sys in training variants of the Criteo \dlrm in MLPerf~\cite{MLPerf} across 20 nodes. \sys recovers from failures significantly faster than checkpointing and with lower training-time overhead. \sys's benefits improve for larger \dlrms, showing promise for current and future \dlrms. For example, \sys reduces training-time overhead for a large \dlrm by up to 88\% compared to checkpointing (from 33.4\% to 4\%).  %
Furthermore, \sys recovers from failure up to 10.3$\times$ faster than the average case for checkpointing, and enables training to continue during recovery with only a 6--12\% drop in throughput, while checkpointing pauses training during recovery. \sys's benefits come at the cost of additional memory requirements and load on the training cluster. However, \sys keeps memory overhead to only a fractional amount and balances additional load evenly among \servers. 
These results show the promise of erasure coding to enable efficient fault tolerance in \dlrm training.

\section{Challenges in fault-tolerant \dlrm training} \label{sec:background}
We next describe \dlrm training systems, the insufficiency of current approaches to fault-tolerant \dlrm training, and opportunity to impart more efficient fault tolerance in \dlrm training.

\subsection{\dlrm training systems} \label{sec:background:dlrm}
As described in \Section\ref{sec:intro}, \dlrms are large in size due to \embeddingTables that span hundreds of gigabytes to terabytes in size, and \dlrm training is typically distributed across a set of \servers and \workers (see \Figure\ref{fig:background:embedding}).  Model parameters are sharded across \server memory. \Workers read \embeddingTableEntries, perform a forward and backward pass over a \nnShort to generate gradients (for both \nnShort parameters and \embeddingTableEntries), and send gradients back to the \servers hosting the \entries. An \optimizer (\eg Adam) on each \server uses gradients received from \workers to \update parameters. Each training sample typically accesses only a few \embeddingTableEntries, but all \nnShort parameters. Thus, \embeddingTableEntries are updated sparsely, while \nnShort parameters are updated on every training sample. Finally, many systems use asynchronous training when training \dlrms (\eg Facebook and Alibaba~\cite{jiang2019xdl,acun2020understanding}). We thus focus on asynchronous training systems in this work, but describe in \Section\ref{sec:true_discussion:sync} how the techniques we propose could apply to synchronous training as well.

Many popular \optimizers use per-parameter state in updating parameters (\eg Adam~\cite{kingma2015adam}, Adagrad~\cite{duchi2011adaptive}, momentum SGD). We refer to such \optimizers as ``stateful \optimizers.'' For example, Adagrad tracks the sum of squared gradients for each parameter over time and uses this when updating the parameter. Per-parameter \optimizer state is kept in memory alongside model parameters on \servers and is updated when the corresponding parameter is updated. As per-parameter state grows with \dlrm size, such \optimizer state for \embeddingTables can consume a large amount of memory.

\subsection{Checkpointing and its downsides} \label{sec:background:checkpointing} \label{sec:background:checkpointing:overhead}
Given the large number of nodes on which \dlrms are trained, failures are common~\cite{maeng2021cpr}. Due to the time it takes to train such models, it is critical that \dlrm training be made fault tolerant. 

We focus on imparting fault tolerance against \server failures. Handling \server failures is critical, as failure of a single \server results in loss of fraction of \embeddingTableEntries. In contrast additional fault tolerance is not needed for handling \worker failures, as each worker contains a replica of the \dlrm's \nn, which can simply be recovered from another worker.

Checkpointing is the primary approach used for fault tolerance in \dlrm training~\cite{maeng2021cpr}. Under checkpointing, training is periodically paused and \dlrm parameters and \optimizer state are writen to stable storage (\eg a distributed file system). Upon failure, the most recent checkpoint is read from stable storage, and the entire system restarts training from this checkpoint, redoing any training iterations that occurred between the most recent checkpoint and the failure.

Recently, \textit{Facebook reported that overheads from checkpointing account for, on average, 12\% of \dlrm training time}, and that these overheads add up to \textit{over 1000 machine-years of computation}~\cite{maeng2021cpr}. 

We next describe two primary time penalties that make up the overhead of checkpointing on training time of \dlrms, and illustrate their impact with varying size of the embedding table and the checkpointing interval.

\begin{figure}[t]
 \centering
  \begin{subfigure}{.49\columnwidth}
    \centering
    \newcommand{\figureheight}{1.12in}
    \newcommand{\figurewidth}{\linewidth}
    \begin{tikzpicture}[]
\pgfplotsset{label style={font=\footnotesize}, 
             tick label style={font=\footnotesize},
             legend style={font=\footnotesize},
             every non boxed x axis/.append style={x axis line style=-},
             every non boxed y axis/.append style={y axis line style=-},
             axis lines=left,
             /pgfplots/ybar legend/.style={
                /pgfplots/legend image code/.code={%
                    \draw[##1,/tikz/.cd,yshift=-0.25em]
                    (0cm,0cm) rectangle (12pt,6pt);
                },
            },
        }

\begin{axis}[
height=\figureheight,
legend cell align={left},
legend columns=1,
legend style={at={(0.5,1.8)}, anchor=north, draw=none, /tikz/every even column/.append style={column sep=0.4cm}},
tick align=outside,
tick label style={/pgf/number format/assume math mode},
tick pos=left,
width=\figurewidth,
x grid style={white!69.01960784313725!black},
xmin=-0.41, xmax=2.71,
xtick={0.15,1.15,2.15},
xticklabels={44, 88, 176},
xlabel style={align=left},
xlabel={Embedding table\\size per server (GB)},
y grid style={gray, opacity=0.3},
ylabel style={at={(axis description cs:0.1,.5)}},
ymajorgrids,
ylabel style={align=left},
ylabel={Time (minutes)\\},
ymin=0, ymax=12
]

\addlegendimage{ybar,ybar legend,draw=black,fill=color_blue_2_1};
\addlegendentry{Write checkpoint}

\addlegendimage{ybar,ybar legend,draw=black,fill=color_blue_2_0};
\addlegendentry{Read checkpoint}

\draw[draw=black,fill=color_blue_2_1] (axis cs:-0.15,0) rectangle (axis cs:0.15,\ckptOgWriteMinutes);
\draw[draw=black,fill=color_blue_2_0] (axis cs:0.15,0) rectangle (axis cs:0.45,\ckptOgReadMinutes);
\draw[draw=black,fill=color_blue_2_1] (axis cs:0.85,0) rectangle (axis cs:1.15,\ckptTwoSparseWriteMinutes);
\draw[draw=black,fill=color_blue_2_0] (axis cs:1.15,0) rectangle (axis cs:1.45,\ckptTwoSparseReadMinutes);
\draw[draw=black,fill=color_blue_2_1] (axis cs:1.85,0) rectangle (axis cs:2.15,\ckptTwoSparseTwoDenseWriteMinutes);
\draw[draw=black,fill=color_blue_2_0] (axis cs:2.15,0) rectangle (axis cs:2.45,\ckptTwoSparseTwoDenseReadMinutes);

\end{axis}

\end{tikzpicture}
    \vspace{-0.2in}
    \caption{Checkpoint write/read time}
    \label{fig:background:checkpointing_time}
  \end{subfigure} \hfill
  \begin{subfigure}{.49\columnwidth}
    \centering
    \newcommand{\figureheight}{1.12in}
    \newcommand{\figurewidth}{\linewidth}
    \begin{tikzpicture}[]
\pgfplotsset{label style={font=\footnotesize}, 
             tick label style={font=\footnotesize},
             legend style={font=\footnotesize},
             every non boxed x axis/.append style={x axis line style=-},
             every non boxed y axis/.append style={y axis line style=-},
             axis lines=left,
             /pgfplots/ybar legend/.style={
                /pgfplots/legend image code/.code={%
                    \draw[##1,/tikz/.cd,yshift=-0.25em]
                    (0cm,0cm) rectangle (12pt,6pt);
                },
            },
        }

\begin{axis}[
height=\figureheight,
legend cell align={left},
legend columns=1,
legend style={at={(0.5,1.8)}, anchor=north, draw=none, /tikz/every even column/.append style={column sep=0.4cm}},
tick align=outside,
tick label style={/pgf/number format/assume math mode},
tick pos=left,
width=\figurewidth,
x grid style={white!69.01960784313725!black},
xmin=-0.41, xmax=2.71,
xtick={0.15,1.15,2.15},
xticklabels={44,88,176},
xlabel style={align=left},
xlabel={Embedding table\\size per server (GB)},
y grid style={gray, opacity=0.3},
ylabel style={at={(axis description cs:0.1,.5)}},
ymajorgrids,
ylabel style={align=left},
ylabel={Increase in\\training time (\%)},
ymin=0, ymax=35
]

\addlegendimage{ybar,ybar legend,draw=black,fill=color_blue_2_1};
\addlegendentry{Ckpt every 30 min.}

\addlegendimage{ybar,ybar legend,draw=black,fill=color_blue_2_0};
\addlegendentry{Ckpt every 60 min.}

\draw[draw=black,fill=color_blue_2_1] (axis cs:-0.15,0) rectangle (axis cs:0.15,\evalNormalCriteoSizeOgCkptEight);
\draw[draw=black,fill=color_blue_2_0] (axis cs:0.15,0) rectangle (axis cs:0.45,\evalNormalCriteoSizeOgCkptFour);
\draw[draw=black,fill=color_blue_2_1] (axis cs:0.85,0) rectangle (axis cs:1.15,\evalNormalCriteoSizeTwoSparseCkptEight);
\draw[draw=black,fill=color_blue_2_0] (axis cs:1.15,0) rectangle (axis cs:1.45,\evalNormalCriteoSizeTwoSparseCkptFour);
\draw[draw=black,fill=color_blue_2_1] (axis cs:1.85,0) rectangle (axis cs:2.15,\evalNormalCriteoSizeTwoSparseTwoDenseCkptEight);
\draw[draw=black,fill=color_blue_2_0] (axis cs:2.15,0) rectangle (axis cs:2.45,\evalNormalCriteoSizeTwoSparseTwoDenseCkptFour);

\end{axis}

\end{tikzpicture}
    \vspace{-0.2in}
    \caption{Training-time overhead}
    \label{fig:background:checkpointing_overhead}
  \end{subfigure}
  \vspace{-0.1in}
  \caption{Time required to read and write checkpoints and overhead of checkpointing on training time with varying \embeddingTable size.}
  \vspace{-0.2in}
\end{figure}

\paragraphOrTextbf{1.\ Time penalty during normal operation.}  
Writing checkpoints to stable storage is a slow process given the large sizes of \embeddingTables and \optimizer state, and training is paused during this time so that the saved model is consistent. Intuitively, the overhead of checkpointing on normal operation increases the more frequently checkpoints are taken and the longer it takes to write a checkpoint (and thus the larger the \dlrm).

To illustrate this overhead, we evaluate checkpointing \dlrms in \xdl. Training is performed on a cluster of 15 \workers and 5 \servers, with checkpoints periodically written to an HDFS cluster. 
We train the \dlrm used for the Criteo Terabyte dataset in MLPerf, which requires 220 GB of memory for \embeddingTables (44 GB per \server). We additionally evaluate with increased \dlrm size by increasing the size of \embeddingTables. The full setup used for this evaluation is described in \Section\ref{sec:evaluation:setup}.

\Figure\ref{fig:background:checkpointing_time} shows that the time overhead for writing checkpoints is significant (on the order of minutes) and increases with increasing \embeddingTable size per \server. \Figure\ref{fig:background:checkpointing_overhead} shows the overhead of checkpointing on training in the absence of failures with two checkpointing periods: 30 and 60 minutes. We measure the time it takes for a each setup to reach the same number of iterations that a system with no fault tolerance (and thus no overhead) reaches in four hours. As expected, training time increases both with increased \dlrm size and with decreased time between checkpoints.

\paragraphOrTextbf{2.\ Time penalty during recovery.} 
Upon failure, a system using checkpointing must roll back the \dlrm to the state of the most recent checkpoint by reading it from stable storage, and redo all training iterations that occurred between this checkpoint and the failure. New training iterations are paused during this time. The time needed to read checkpoints from storage can be significant~\cite{maeng2021cpr} and grows with \dlrm size. The expected time to redo iterations grows with the time between checkpoints: if checkpoints are written every $\checkpointPeriod$ time units, this time will be 0 at best (failing just after writing a checkpoint),  $\checkpointPeriod$ at worst (failing just before writing a checkpoint), and $\frac{\checkpointPeriod}{2}$ on average.

\paragraphOrTextbf{Takeaway.} Checkpointing suffers a fundamental tradeoff between training-time overhead in normal operation and when recovering from failure. Increasing the time between checkpoints reduces the fraction of time paused when saving checkpoints, but increases the expected work to be redone in recovery. Furthermore, these overheads increase with model size. Given the trend of increasing model size~\cite{huang2020mixed} \textit{checkpointing is slated to become an even larger overhead in training future \dlrms.} This calls for alternate approaches to fault tolerance in \dlrm training.

\subsection{Reducing overhead via approximation?} \label{sec:background:approximate}
{
Several recent approaches aim to reduce checkpointing overhead by taking approximate checkpoints or via approximate recovery~\cite{qiao2019fault,chen2020efficient,eisenman2020check,maeng2021cpr}. However, in the event of a failure, such techniques roll back an approximation of the true \dlrm, which can potentially alter convergence and final accuracy. Given the significant business value generated by \dlrms, prior works~\cite{zhao2019aibox,eisenman2020check} have noted that even small drops in \dlrm accuracy must be avoided. Furthermore, our personal conversations with {multiple practitioners working on large-scale \dlrm training} 
indicate that this potential accuracy drop introduces a source of uncertainty that makes debugging production systems difficult, and thus is less desirable.
{Hence, ideally, one would like to reduce the overhead of checkpointing without compromising accuracy.} 

Another approach to reducing the overhead of checkpointing is to asynchronously write checkpoints while training progresses by writing updates to stable storage as they are generated~\cite{abadi2016tensorflow}. This is feasible only if writing to stable storage can keep pace with the rate at which gradients are generated. As \dlrm training systems have many workers generating gradients asynchronously, gradients are generated at a high rate that stable storage cannot keep pace with. In fact, if storage could keep pace, then \embeddingTables could be kept in stable storage, rather than in memory. Thus, asynchronous checkpointing is not a viable alternative for fault-tolerant \dlrm training.}

\subsection{Fault tolerance via  in-memory redundancy?} \label{sec:background:redundancy}
An alternative to checkpointing is to provision extra memory in the system to \textit{redundantly} store \dlrm parameters and optimizer state in memory in a fault-tolerant manner.

\paragraphOrTextbf{Replication.} The simplest and most common approach to redundancy is replication, in which a system proactively provisions redundant \servers that contain copies of the \dlrm parameters that are kept up-to-date throughout training and that can immediately take over for failed \servers. Replicated \dlrm training would use twice as much memory to store copies of each \dlrm parameter on two \servers. Gradients for a given parameter are sent to and applied on both \servers holding copies. The system seamlessly continues training if a single \server fails by accessing the replica, avoiding the need of checkpointing-based approaches to redo work after a failure. Similar to traditional checkpointing-based approaches, replicated \dlrm training would preserve the accuracy guarantees of the underlying training system. However, a replicated \dlrm training system requires at least twice as much memory as a non-replicated one. Given the large sizes of \embeddingTables and \optimizer state, this memory overhead is impractical. %

\paragraphOrTextbf{Erasure codes.} \label{sec:opportunity:ec}
Erasure codes are coding-theoretic tools that enable redundancy with low overhead. They have been used for imparting resilience against unavailability in storage and communication systems with significantly less overhead than replication~\cite{patterson1988raid,rizzo1997effective,weatherspoon2002erasure}. An erasure code encodes $k$ data units to generate $r$ redundant ``parity units’’ such that any $k$ out of the total $(k + r)$ data and parity units suffice for a decoder to recover the original $k$ data units. Erasure codes operate with overhead of $\frac{k+r}{k}$, which is less than that of replication by setting $r < k$.  These properties have led to wide adoption of erasure codes in storage, communication, and caching systems~\cite{patterson1988raid,rizzo1997effective,weatherspoon2002erasure,rashmi2016eccache,yan2017tiny}.

\newcommand{\exData}{x\xspace}
\newcommand{\exDataOne}{\exData_1\xspace}
\newcommand{\exDataTwo}{\exData_2\xspace}
\newcommand{\exDataThree}{\exData_3\xspace}
\newcommand{\exParity}{p\xspace}
For example, consider an erasure-coded storage system in which data units $\exDataOne$, $\exDataTwo$, and $\exDataThree$ are stored on three separate disks, and that the system must tolerate one disk failure. An erasure code with parameters $k=3$ and $r=1$ would do so by encoding a parity unit as $\exParity = \exDataOne + \exDataTwo + \exDataThree$ and storing this parity unit on a fourth disk. Suppose the disk holding $\exDataTwo$ fails. The system recovers $\exDataTwo$ using the erasure code's subtraction decoder: $\exDataTwo = \exParity - \exDataOne - \exDataThree$. This setup can recover from any one of the four disks failing by using only one extra disk, while replication would require three extra disks to impart the same level of fault tolerance.

\Figure\ref{fig:background:embedding_coded} shows an example of how erasure codes might potentially be used in \dlrm training to  reduce the memory overhead of in-memory redundancy in \dlrm training. However, there are several challenges in using erasure codes for \dlrm training, which we discuss and address in the remainder of the paper.

\paragraphOrTextbf{Takeaway.} An ideal approach to fault-tolerant \dlrm training would have (1) low-latency recovery, (2) low memory overhead, (3) no potential for accuracy loss. Erasure codes offer promising potential for achieving these goals. 
{However, there are several challenges in using erasure codes for \dlrm training. We describe these challenges in detail and how they can be overcome in the next section.}

\section{\sys: erasure-coded \dlrm training} \label{sec:opportunity}
We propose \sys, a system that imparts efficient fault tolerance to \dlrm training via careful design adapting simple erasure codes and ideas from storage systems.  
{\sys makes the same accuracy guarantees as the underlying training system, unlike the approximate approaches described in \Section\ref{sec:background:approximate}}. 

Using erasure codes in \dlrm training raises unique challenges compared to the traditional use of erasure codes in storage and communication systems. We first provide a high-level overview of \sys and then introduce these challenges and how \sys overcomes them.

\subsection{Overview of \sys} \label{sec:opportunity:architecture}
\Figure\ref{fig:design:update_server} shows the high-level operation of \sys. \sys encodes \dlrm parameters using an erasure code and distributes parities throughout the cluster before training begins. Groups of $k$ parameters from separate \servers are encoded to produce $r$ parities that are placed in memory on separate \servers. \sys thus uses $\frac{k+r}{k}$-times as much memory as the original system. We describe in \Section\ref{sec:design:coding} which \dlrm parameters \sys encodes and how parities are placed in the cluster. As \dlrm parameters are updated during training, \sys updates the corresponding parities. When a \server fails, \sys uses the erasure code's decoder to reconstruct lost parameters.

While the use of erasure codes in \dlrm training is enticing, it presents many challenges and design decisions: (1) Which parameters should be encoded and where should parities be placed (\Section\ref{sec:design:coding})? (2) How can parities be updated correctly and efficiently (\Section\ref{sec:technique:server})? (3) How can \sys avoid pausing training during recovery (\Section\ref{sec:technique:recovery})? (4) How can \sys guarantee the consistency of the \dlrm recovered after failure (\Section\ref{sec:discussion:consistency})?
We next describe how \sys addresses these challenges. %

\subsection{Encoding and placing parity parameters} \label{sec:design:coding}
{We next describe which parameters of a \dlrm are encoded in \sys and where parities are placed.} %

\paragraphOrTextbf{Which parameters should be encoded?} Fault tolerance is primarily needed in \dlrm training to recover failed \servers, which hold \dlrm parameters and \optimizer state. If a \server fails, the portion of the \dlrm hosted on that server is lost, and training cannot proceed. In contrast, systems with architectures as described in \Section\ref{sec:background:dlrm} are naturally tolerant of \worker failures, as the system can continue training with fewer \workers while replacements are provisioned (albeit, at lower training throughput).

As each \worker pulls all \nnShort parameters from \servers when training, the \nnShort in the \dlrm is naturally replicated on \workers. If a \server fails, the \nnShort parameters it held can be recovered from a \worker.\footnote{While  asynchronous training does not guarantee that all \workers have up-to-date \nnShort parameters, the \nnShort recovered from a \worker is equivalent to one observable under asynchronous training.} 

In contrast, \embeddingTables and \optimizer state are \textit{not} naturally replicated. \EmbeddingTables and \optimizer state are sharded across \servers, and each \worker reads only a few \entries each training iteration. Thus, lost \embeddingTableEntries and \optimizer state cannot be recovered from \workers. Furthermore, replicating \embeddingTables and \optimizer state is impractical, given their large size.

Thus, \sys encodes only \embeddingTables and \optimizer state, while \nnShort parameters need not be encoded.

\begin{figure}[t]
    \centering
    \includegraphics[width=0.85\linewidth]{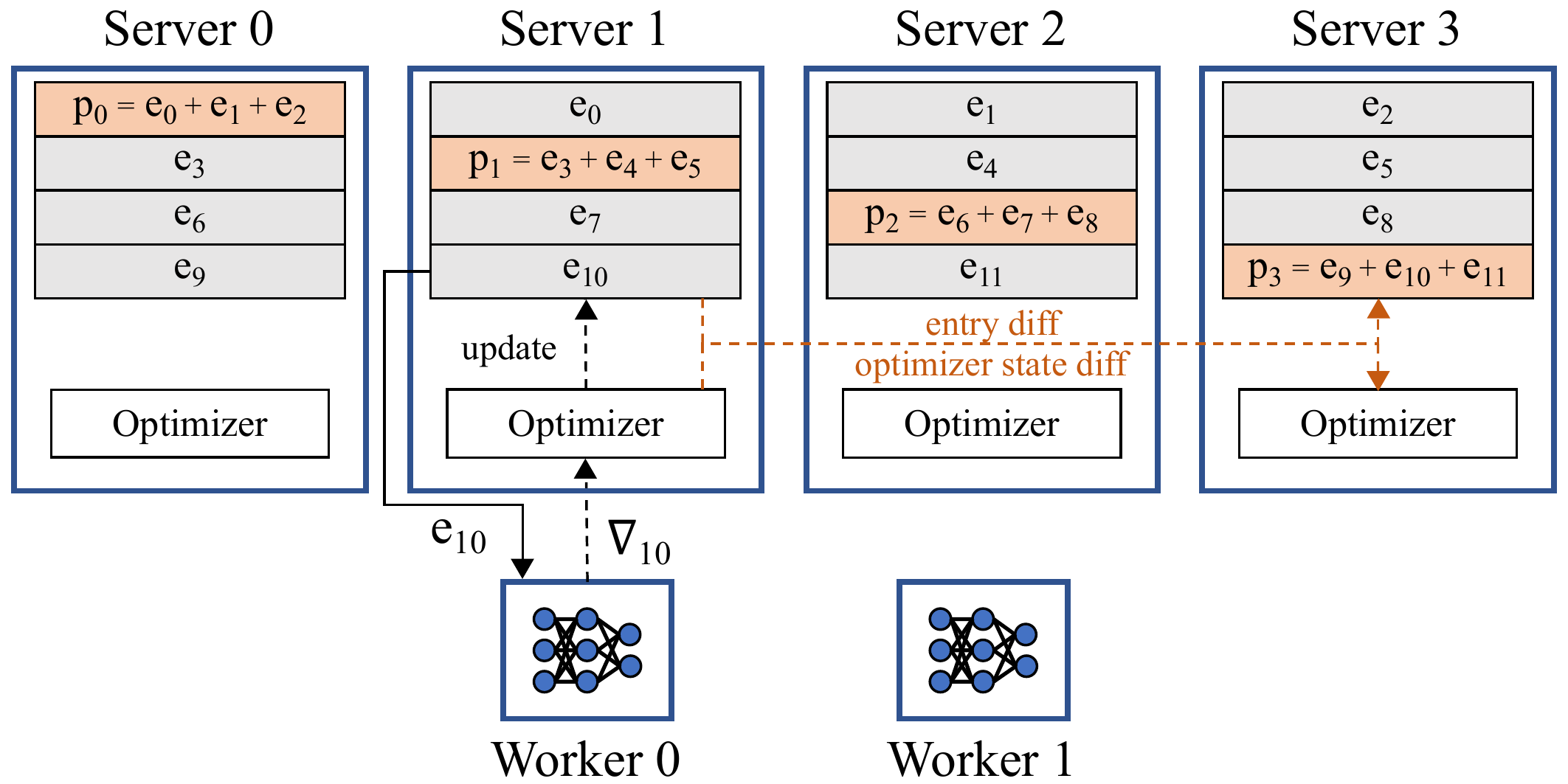}
    \vspace{-0.1in}
    \caption{Example of \sys with $k=3$, $r=1$.}
    \label{fig:design:update_server}
\end{figure}

\paragraphOrTextbf{Where should parities be placed?}
Recall from \Section\ref{sec:background:dlrm} that \embeddingTables and \optimizer state are sharded across \servers. \sys encodes groups of $k$ \embeddingTableEntries from different shards to produce a ``\parityEntry,'' and places the \parityEntry on a separate \server. \Optimizer state is similarly encoded to form ``parity \optimizer state,'' which is placed on the same \server hosting the corresponding \parityEntry.

Parities in \sys are updated whenever any of the $k$ corresponding \embeddingTableEntries are updated. Hence, parities are updated more frequently than the original \entries, and must be placed carefully within the cluster so as not to introduce load imbalance among \servers. \sys uses rotating parity placement to distribute parities among \servers, resulting in an equal number of parities per \server. An example of this is shown in \Figure\ref{fig:design:update_server} with $k=3$: each \server is chosen to host a parity in a rotating fashion, and the \entries used to encode the parity are hosted on the 3 other \servers. This approach is inspired by parity placement in RAID-5 hard-disk systems~\cite{patterson1988raid}.

\paragraphOrTextbf{Encoder and decoder.} We focus on using erasure codes with parameter $r = 1$ (\ie one parity per $k$ \entries, and recovering from a single failure). We focus on this setting because it represents the most common failure scenario experienced by a cluster in datacenters~\cite{rashmi2014hitchhiker}. {We describe in \Section\ref{sec:true_discussion:concurrent} how \sys can be easily adapted to handle concurrent failures with $r > 1$.} Within this setting of $r=1$, \sys uses the simple summation encoder shown in \Figure\ref{fig:design:update_server}, and the corresponding subtraction decoder. For example, with $k=3$, \embeddingTableEntries 
$\embOne$, $\embTwo$, and $\embThree$ are encoded to form parity $\parity = \embOne + \embTwo + \embThree$. If the \server holding $\embTwo$ fails, $\embTwo$ will be recovered as $\embTwo = \parity - \embOne - \embThree$.

\subsection{Correctly and efficiently updating parities} \label{sec:technique:server}
We next describe challenges in correctly and efficiently updating parities, and how \sys overcomes them.

\subsubsection{Challenges in keeping up-to-date parities} \label{sec:technique:server:challenge}
\paragraphOrTextbf{Maintaining correctness with stateful \optimizers.} \label{sec:challenge:optimizer}
The naive approach to erasure-coded \dlrm training shown in \Figure\ref{fig:background:embedding_coded} suffers a fundamental challenge in \textit{correctly} updating \parityEntries when using a stateful \optimizer. 

Consider the example in \Figure\ref{fig:background:embedding_coded} when using the Adagrad \optimizer. Let $\embIT$ denote the value of \embeddingTableEntry $\embI$ after $\myTime$ updates, and $\gradIT$ denote the gradient for $\embIT$. The \update performed by Adagrad for $\embOneT$ with gradient $\gradOneT$ is:
\begin{equation*}
    \embOneTPlusOne = \embOneT - \frac{\lr}{\sqrt{\adagradAccumOneT + \adagradEps}}\gradOneT
\end{equation*}
where $\lr$ is a constant learning rate, $\adagradAccumOneT = \gradOneTZero^2 + \gradOneTOne^2 + \hdots + \gradOneT^2$ is the sum of squares of the previous gradients for parameter $\embOne$, and $\adagradEps$ is a small constant. $\adagradAccumOneT$, which we call $\embOne$'s ``\accumulator,'' is an example of per-parameter \optimizer state.

As described in \Section\ref{sec:opportunity:architecture}, \sys maintains one ``parity optimizer parameter'' for every $k$ original optimizer parameters. In the example above, using the encoder described in \Section\ref{sec:opportunity:architecture}, a ``\parityAccumulator'' would be $\adagradAccumP = \adagradAccumOne + \adagradAccumTwo + \adagradAccumThree$. This is easily kept up-to-date by adding to the \parityAccumulator the squared gradients for updates to each of the $k$ original \embeddingTableEntries. However, using such a \parityAccumulator to update the corresponding \parityEntry based on $\gradOneT$ would result in an incorrect \parityEntry, as $\adagradAccumOneT \neq \adagradAccumPT$. 

The issue illustrated in the example above arises for any stateful \optimizer, such as Adagrad, Adam, and momentum SGD. Given the popularity of such \optimizers, \sys must employ some means of maintaining correct parities when using stateful \optimizers. This issue could be overcome by keeping replicas the $k$ original \optimizer parameters on the \server hosting the parity. However, as described in \Section\ref{sec:opportunity:architecture}, \optimizer state is large and grows with \embeddingTables, so such replication is impractical. 

\paragraphOrTextbf{Maintaining low overhead.} \label{sec:challenge:normal} 
Even if the issue above were not present, the approach to erasure-coded \dlrm training shown in \Figure\ref{fig:background:embedding_coded} can have high training-time overhead. Under this approach, updating parities requires gradients for an \entry to be sent both to the \server hosting the \entry and to the \server hosting the parity, and that the \optimizer be applied on both servers. This results in network and compute overhead for \workers. {Given that \workers are typically the bottleneck in \dlrm training~\cite{jiang2019xdl}, }\sys must minimize this overhead. 

\paragraphOrTextbf{Fundamental limitation underlying the challenges.} The challenges described above stem from sending gradients directly to the \servers hosting parities, an approach we call ``\workerSideProp.'' 
Under \workerSideProp, \workers must send duplicate gradients, resulting in CPU and network overhead on \workers. \Servers holding \parityEntries receive only the gradient for the original \embeddingTableEntry and must correctly \update the \parityEntry and \optimizer state. As described above, performing these updates correctly given only gradients and parity \optimizer state is challenging. 

\subsubsection{\ServerSideProp} \label{sec:technique:server:server}
To overcome these challenges, \sys introduces \textit{\serverSideProp}. As shown in \Figure\ref{fig:design:update_server}, under \serverSideProp, \workers send gradients only to the \servers holding \embeddingTableEntries for that gradient. After applying the \optimizer to \embeddingTableEntries and updating \optimizer state, the \server asynchronously sends the \textit{differences} in \entry and \optimizer state to the \server holding the corresponding \parityEntry. The receiving \server adds these differences to the \parityEntry and \optimizer state.

\ServerSideProp has two key benefits over \workerSideProp. (1) By sending differences to \servers, rather than gradients, \serverSideProp updates \parityEntries correctly when using stateful \optimizers. (2) \ServerSideProp adds no overhead to \workers, which are often the bottleneck in \dlrm training~\cite{jiang2019xdl}. 

While \serverSideProp does introduce network and CPU overhead on \servers for sending and applying differences, \Section\ref{sec:evaluation} will show that it significantly outperforms \workerSideProp. {As an additional note, the differences sent by \serverSideProp can be dense (\eg due to momentum used by the \optimizer), and thus less amenable to sparsity-based compression than gradients. However, \dlrm training clusters typically have high-bandwidth networks (\eg 100 Gbps~\cite{acun2020understanding}), for which compression has been shown to provide little benefit~\cite{zhang2020network}.}

\subsection{Pause-free recovery from failure} \label{sec:technique:recovery}
We next describe how \sys recovers from failure without requiring training to pause.

\paragraphOrTextbf{Challenges in erasure-coded recovery.}  \label{sec:challenge:recovery}
Due to the property of the erasure codes described in \Section\ref{sec:background:redundancy} that any $k$ out of the $(k + 1)$ original and parity units suffice to recover the original $k$ units, \sys can continue training even if a single \server fails. For example, a \worker in \sys could read \entry $\embTwo$ in \Figure\ref{fig:design:update_server} even if \Server~2 fails by reading $\embOne$, $\embThree$, and $\parity$, and decoding $\embTwo = \parity - \embOne - \embThree$. Such read operations that require decoding are referred to as ``degraded reads'' in erasure-coded storage systems.

Despite the ability to perform degraded reads, \sys must still fully recover failed \servers to remain tolerant of future failures. However, prior work on erasure-coded storage has shown that full recovery can be time-intensive~\cite{sathiamoorthy2013xoring,rashmi2014hitchhiker}. Full recovery in \sys requires decoding all \embeddingTableEntries and \optimizer state held by the failed \server. This consumes significant network bandwidth in transferring available \entries for decoding, and \server CPU in performing decoding. Given the large sizes \embeddingTableEntries and \optimizer state, completing full recovery before resuming training can significantly pause training. 

\paragraphOrTextbf{Training during recovery in \sys.} \label{sec:technique:recovery:nonblocking}
Rather than solely performing degraded reads after a failure or pausing until full recovery is complete, \sys \textit{enables training to continue while full recovery takes place.} Upon failure, \sys begins full recovery of lost \embeddingTableEntries and \optimizer state. In the meantime, the system continues to perform new training iterations, with \workers performing degraded reads to access \entries from the failed \server. 

\sys must avoid updating an \embeddingTableEntry in parallel with its use for recovery. If the recovery process reads the new value of the \entry, but the old value of the \parityEntry (\eg because the update has not yet reached the parity), then the recovered \entry will be incorrect. \sys uses granular locking to avoid such race conditions. The recovery process ``locks'' a fraction of the lost \entries that it will decode. While this lock is held, all \updates to \entries that will be used in recovery for the locked chunk are buffered in memory on \servers. \Workers reading an updated, but locked \entry do so by reading from the buffer. When a lock is released, all buffered \updates are applied to the \embeddingTables, and the  next chunk is locked. The number of \entries covered by each lock introduces a tradeoff between time overhead in switching locks and \server memory overhead for buffering that can be navigated based on the requirements of a given system. 

\subsection{Consistency of recovered \dlrm}\label{sec:discussion:consistency}
\sys provides the same guarantees regarding the consistency of a recovered \dlrm as the general asynchronous system it builds atop.

\paragraphOrTextbf{Consistency of each parameter.} \sys ensures that each \embeddingTableEntry and \optimizer parameter is recovered to the value corresponding to its most recent update that was applied both to the original \entry and the parity. One case that requires care: if recovery is triggered after an update has been applied to an \embeddingTableEntry but before it is applied to the corresponding \parityEntry, the decoded \entry will be incorrect. \sys avoids this scenario by ensuring that all in-flight \updates are completed before recovery begins. 

\paragraphOrTextbf{Consistency across parameters.} \sys ensures that the recovered parameters, taken as a whole, represent a \dlrm that could have been achieved by asynchronous training. \sys cannot guarantee that the recovered \dlrm represents an exact state observed in training. However, we show in \Section\ref{sec:appendix:consistency} of the appendix that, when such scenarios occur, the recovered \dlrm is equivalent to one that could have occurred during asynchronous training. Thus, \sys does not introduce additional inconsistency atop general asynchronous training.

\subsection{Tradeoffs in \sys} \label{sec:discussion} \label{sec:discussion:k}
\sys encodes $k$ \embeddingTableEntries into a single \parityEntry ($r=1$) (similarly for \optimizer state). Parameter $k$ results in the following tradeoffs in \sys, some of which differ from those in the traditional use of erasure codes:

\paragraphOrTextbf{Increasing $\mathbf{k}$ decreases memory overhead and fault tolerance.} As \sys encodes one \parityEntry for every $k$ \embeddingTableEntries (and similarly for \optimizer state), \sys requires less memory for storing parities with increased $k$. However, since the erasure codes employed by \sys can recover from any one out of $(k + 1)$ failures, increasing $k$ decreases the fraction of failed \servers \sys can tolerate. 

\newcommand{\exEmbEntries}{e\xspace}
\newcommand{\exServers}{s\xspace}
\newcommand{\exNumUpdate}{u\xspace}
\paragraphOrTextbf{Increasing $\mathbf{k}$ \textit{does not} change load during normal operation.} As each \embeddingTableEntry in \sys is encoded to produce a single \parityEntry, each update applied to an \entry is also be applied to one parity. Thus, the total increase in load due to \sys is $2\times$, regardless of the value of $k$. In addition to this constant load increase, we show in \Section\ref{sec:evaluation:results:normal} that \sys balances this load evenly with various values of $k$.

\paragraphOrTextbf{Increasing $\mathbf{k}$ increases the time to fully recover.} Recovery in \sys requires reading $k$ available \entries from separate \servers and decoding. Thus, the network traffic and computation used during recovery increases with $k$, which increases the time to fully recover a failed \server. However, as described in \Section\ref{sec:technique:recovery:nonblocking}, \sys allows training to continue during this time. 

\section{Evaluation} \label{sec:evaluation}
We next evaluate the performance of \sys. The highlights of the evaluation are as follows:
\begin{denseitemize}
\item \sys recovers from failure up to 10.3$\times$ faster than the average recovery time for checkpointing. 
\item \sys enables training to proceed during recovery with only a 6\%--12\% drop in throughput, whereas checkpointing requires training to completely pause. 
\item \sys reduces training-time overhead by up to 88\% compared to checkpointing (from 33.4\% to 4\%) on large \dlrms.  
\item While \sys introduces additional load for updating parities, the impact of this increased load on training throughput is alleviated by improved cluster load balance.
\item \sys operates with low training-time overhead in a variety of cluster configurations and recovers quickly from failure with varying lock granularity (see \Section\ref{sec:technique:recovery:nonblocking}).
\end{denseitemize}

\subsection{Evaluation setup} 
\label{sec:evaluation:setup}
We implement \sys in C++ on XDL, an open-source \dlrm training system from Alibaba~\cite{jiang2019xdl}.

\paragraphOrTextbf{Dataset.} We evaluate with the Criteo Terabyte dataset~\cite{Criteo}. We randomly draw one day of samples from the dataset by picking each sample with probability $\frac{1}{24}$ in one pass through the dataset, and use this subset in evaluation to reduce storage requirements. This random sampling results in a sampled dataset that mimics the full dataset.

\paragraphOrTextbf{Models.} We use the \dlrm for the \criteo dataset from MLPerf~\cite{naumov2019deep}, which has 13 \embeddingTables, for a total of nearly 200M \entries each with 128 dense features. We use SGD with momentum as the \optimizer, which add one floating point value of \optimizer state per parameter. Any other \optimizer can also be used. The total size of the \embeddingTables and \optimizer state is \criteoOgSizeGB. The \dlrm uses a seven-layer multilayer perceptron with 128--1024 features per layer as a \nnShort~\cite{MLPerf-inference}.

We evaluate \dlrms of different size by varying \embeddingTable size in two ways: (1) Increasing the number of \entries (\ie sparse dimension). This increases the memory required per \server and the amount of data that must be checkpointed/coded and recovered. (2) Increasing the size of each \entry (\ie dense dimension). This increases the memory required per \server, the amount of data that must be checkpointed/coded, the network bandwidth in transferring \entries/gradients, the work performed by \nnsShort, and the work done by \servers in updating \entries. %
We consider three variants of the \dlrm: (1) \criteoOg, the original \criteo \dlrm described above, (2) \criteoTwoSparse, which has 2$\times$ the \embeddingTableEntries (\ie $2\times$ sparse dimension), and (3) \criteoTwoSparseTwoDense, which has 2$\times$ the number of \entries and with each entry being 2$\times$ as large (\ie $2\times$ sparse and dense dimensions). For \criteoTwoSparseTwoDense, the input layer of the \nnShort is modified to accommodate the larger \entry size.  These variants have size \criteoOgSize, \criteoTwoSparseSize, and \criteoTwoSparseTwoDenseSizeGB, respectively. The two larger \dlrms reflect performance for future \dlrms, which are expected to grow in size~\cite{huang2020mixed}

\newcommand{\progressModel}{\criteoTwoSparseTwoDense}

\begin{figure}[t]
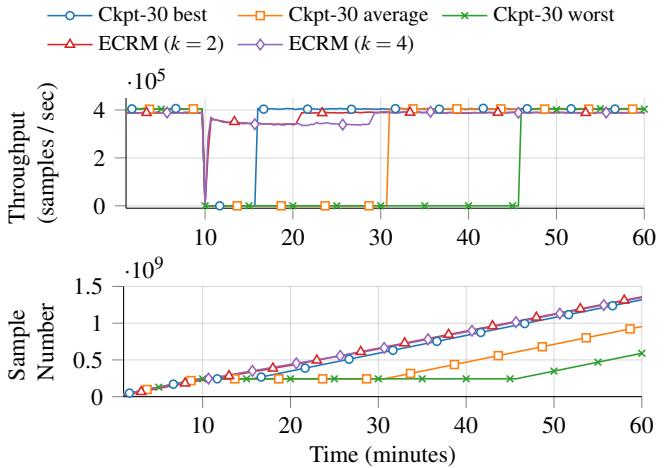

    \centering
    \newcommand{\figureheight}{1.2in}
    \newcommand{\figurewidth}{\linewidth}
        \newcommand{\recoveryTputYmax}{450000}

\begin{tikzpicture}[spy using outlines=
	{rectangle, magnification=3, width=2cm, height=0.7cm, connect spies, rounded corners}]

\pgfplotsset{label style={font=\small}, 
             tick label style={font=\small},
             legend style={font=\footnotesize},
             every non boxed x axis/.append style={x axis line style=-},
             every non boxed y axis/.append style={y axis line style=-},
             axis lines=left}

\definecolor{color0}{rgb}{0.12156862745098,0.466666666666667,0.705882352941177}
\definecolor{color1}{rgb}{1,0.498039215686275,0.0549019607843137}
\definecolor{color2}{rgb}{0.172549019607843,0.627450980392157,0.172549019607843}
\definecolor{color3}{rgb}{0.83921568627451,0.152941176470588,0.156862745098039}
\definecolor{color4}{rgb}{0.580392156862745,0.403921568627451,0.741176470588235}
\definecolor{color5}{rgb}{0.549019607843137,0.337254901960784,0.294117647058824}

\begin{axis}[
height=\figureheight,
legend cell align={left},
legend columns=3,
legend style={at={(0.4,1.3)}, anchor=south, draw=none,
            /tikz/every even column/.append style={column sep=0.1cm}},
tick align=outside,
tick label style={/pgf/number format/assume math mode},
tick pos=left,
width=\figurewidth,
x grid style={gray, opacity=0.3},
xmajorgrids,
xmin=1, xmax=60,%
y grid style={gray, opacity=0.3},
ymajorgrids,
ylabel style={align=left},
ylabel={Throughput\\(samples / sec)},
ymin=-10000, ymax=\recoveryTputYmax
]

        \input{figures/evaluation/data/recovery/criteo-4x/tput}

\end{axis}

\end{tikzpicture}
    \renewcommand{\figureheight}{1.2in}
    \renewcommand{\figurewidth}{\linewidth}
        \newcommand{\recoveryProgressYmax}{1500000000}

\begin{tikzpicture}[spy using outlines=
	{rectangle, magnification=2, width=1cm, height=1cm, connect spies, rounded corners}]

\pgfplotsset{label style={font=\small}, 
             tick label style={font=\small},
             legend style={font=\footnotesize},
             every non boxed x axis/.append style={x axis line style=-},
             every non boxed y axis/.append style={y axis line style=-},
             axis lines=left}

\definecolor{color0}{rgb}{0.12156862745098,0.466666666666667,0.705882352941177}
\definecolor{color1}{rgb}{1,0.498039215686275,0.0549019607843137}
\definecolor{color2}{rgb}{0.172549019607843,0.627450980392157,0.172549019607843}
\definecolor{color3}{rgb}{0.83921568627451,0.152941176470588,0.156862745098039}
\definecolor{color4}{rgb}{0.580392156862745,0.403921568627451,0.741176470588235}
\definecolor{color5}{rgb}{0.549019607843137,0.337254901960784,0.294117647058824}

\begin{axis}[
height=\figureheight,
legend cell align={left},
legend columns=3,
legend style={at={(0.4,1.3)}, anchor=south, draw=none,
            /tikz/every even column/.append style={column sep=0.1cm}},
tick align=outside,
tick label style={/pgf/number format/assume math mode},
tick pos=left,
width=\figurewidth,
x grid style={gray, opacity=0.3},
xmajorgrids,
xmin=1, xmax=60,%
xlabel={Time (minutes)},
y grid style={gray, opacity=0.3},
ymajorgrids,
ylabel style={align=left},
ylabel={Sample\\Number},
ymin=0, ymax=\recoveryProgressYmax%
]
        \input{figures/evaluation/data/recovery/criteo-4x/progress}

\end{axis}

\end{tikzpicture}
    \caption{Training throughput (top) and progress (bottom) when recovering from failure at 10 minutes.}
    \label{fig:eval:recovery_tput}
    \label{fig:eval:recovery_progress}
\end{figure}

\paragraphOrTextbf{Coding parameters and baselines.} We evaluate \sys with $k$ of 2, 4, and 10, which have 50\%, 25\%, and 10\% memory overhead, respectively. We evaluate with $k=10$ in a limited set of experiments due to the cost of the larger cluster needed. {We use one lock during recovery by default (see \Section\ref{sec:challenge:recovery}), but also evaluate other locking granularities.}

We compare \sys to taking checkpoints to HDFS every 30 minutes (\ckptThirty) and every 60 minutes (\ckptSixty). {Checkpointing to HDFS is representative of production \dlrm training environments, which leverage HDFS-like distributed file systems~\cite{acun2020understanding}. Furthermore, the checkpointing baselines we use have competitive performance: we find that checkpointing via HDFS is only 7\%--27\% slower than a (purposely unrealistic) baseline of writing directly to a local SSD. In addition, for the \criteoOg \dlrm, which is representative of current \dlrms, the checkpoint-writing overhead we report in \Section\ref{sec:evaluation:results:normal} is similar to that reported in production training jobs by Facebook~\cite{maeng2021cpr}.} We compare \sys only to approaches to fault tolerance that, like \sys, maintain the same accuracy guarantees as the underlying training system.

\paragraphOrTextbf{Cluster setup.} We evaluate on AWS with 5 \servers of type r5n.8xlarge, each with 32 vCPUs, 256 GB of memory, and 25 Gbps network bandwidth (r5n.12xlarge is used for \criteoTwoSparseTwoDense due to memory requirements). We use 15 \workers of type \expWorkerType, each with a V100 GPU, 8 vCPUs, and 10 Gbps of network bandwidth. This ratio of \worker to \server nodes is inspired by XDL~\cite{jiang2019xdl}. \Workers use batch size of 2048. We consider a varying number of \workers and \servers with limited CPU and network resources in \Section\ref{sec:evaluation:results:normal}. For checkpointing, we use 15 additional nodes of type i3en.xlarge as HDFS nodes, each equipped with NVMe SSDs and 25 Gbps of network bandwidth.  All instances we consider use AWS ENA networking.

\paragraphOrTextbf{Metrics.} For performance during recovery, we measure the time to fully recover a failed server and training throughput during recovery (samples/second). For performance during normal operation, we measure training throughput and training-time overhead, which is the percent increase in the time to train a certain number of samples.

\begin{figure}[t]
    \centering
    \newcommand{\figureheight}{1.2in}
    \newcommand{\figurewidth}{\linewidth}
    \begin{tikzpicture}[]
    \tikzset{
        hatch distance/.store in=\hatchdistance,
        hatch distance=5pt,
        hatch thickness/.store in=\hatchthickness,
        hatch thickness=0.5pt
    }
\pgfplotsset{label style={font=\footnotesize}, 
             tick label style={font=\footnotesize},
             legend style={font=\footnotesize},
             every non boxed x axis/.append style={x axis line style=-},
             every non boxed y axis/.append style={y axis line style=-},
             axis lines=left,
             /pgfplots/ybar legend/.style={
                /pgfplots/legend image code/.code={%
                    \draw[##1,/tikz/.cd,yshift=-0.25em]
                    (0cm,0cm) rectangle (12pt,6pt);
                },
            },
        }
        
            \makeatletter
    \pgfdeclarepatternformonly[\hatchdistance,\hatchthickness]{flexible hatch}
    {\pgfqpoint{0pt}{0pt}}
    {\pgfqpoint{\hatchdistance}{\hatchdistance}}
    {\pgfpoint{\hatchdistance-1pt}{\hatchdistance-1pt}}%
    {
        \pgfsetcolor{\tikz@pattern@color}
        \pgfsetlinewidth{\hatchthickness}
        \pgfpathmoveto{\pgfqpoint{0pt}{0pt}}
        \pgfpathlineto{\pgfqpoint{\hatchdistance}{\hatchdistance}}
        \pgfusepath{stroke}
    }
    \makeatother

\begin{axis}[
height=\figureheight,
legend cell align={left},
legend columns=3,
legend style={at={(0.4,2.1)}, anchor=north, draw=none, /tikz/every even column/.append style={column sep=0.4cm}},
tick align=outside,
tick label style={/pgf/number format/assume math mode},
tick pos=left,
width=\figurewidth,
x grid style={white!69.01960784313725!black},
xmin=-0.172, xmax=2.732,
xtick={0.28,1.28,2.28},
xticklabel style={align=center},
xticklabels={{\criteoOg\\\criteoOgSizeGB},{\criteoTwoSparse\\\criteoTwoSparseSizeGB},{\criteoTwoSparseTwoDense\\\criteoTwoSparseTwoDenseSizeGB}},
y grid style={gray, opacity=0.3},
ylabel style={at={(axis description cs:0.02,.5)}},
ymajorgrids,
ylabel style={align=left},
ylabel={Recovery time\\(minutes)},
ymin=0, ymax=70
]

\addlegendimage{ybar,ybar legend,draw=black,fill=color_red_3_0,postaction={pattern=flexible hatch, pattern color=black}};
\addlegendentry{\ckptSixty best}
\addlegendimage{ybar,ybar legend,draw=black,fill=color_red_3_1,postaction={pattern=flexible hatch, pattern color=black}};
\addlegendentry{\ckptSixty average}
\addlegendimage{ybar,ybar legend,draw=black,fill=color_red_3_2,postaction={pattern=flexible hatch, pattern color=black}};
\addlegendentry{\ckptSixty worst}
\addlegendimage{ybar,ybar legend,draw=black,fill=color_green_3_0,postaction={pattern=north west lines, pattern color=black}};
\addlegendentry{\ckptThirty best}
\addlegendimage{ybar,ybar legend,draw=black,fill=color_green_3_1,postaction={pattern=north west lines, pattern color=black}};
\addlegendentry{\ckptThirty average}
\addlegendimage{ybar,ybar legend,draw=black,fill=color_green_3_2,postaction={pattern=north west lines, pattern color=black}};
\addlegendentry{\ckptThirty worst}
\addlegendimage{ybar,ybar legend,draw=black,fill=color_blue_2_0};
\addlegendentry{\sysKFour}
\addlegendimage{ybar,ybar legend,draw=black,fill=color_blue_2_1};
\addlegendentry{\sysKTwo}

\draw[draw=black,fill=color_red_3_0,postaction={pattern=flexible hatch, pattern color=black}] (axis cs:-0.04,0) rectangle (axis cs:0.04,\evalRecoveryCriteoSizeOgCkptFourBest);
\draw[draw=black,fill=color_red_3_0,postaction={pattern=flexible hatch, pattern color=black}] (axis cs:0.96,0) rectangle (axis cs:1.04,\evalRecoveryCriteoSizeTwoSparseCkptFourBest);
\draw[draw=black,fill=color_red_3_0,postaction={pattern=flexible hatch, pattern color=black}] (axis cs:1.96,0) rectangle (axis cs:2.04,\evalRecoveryCriteoSizeTwoSparseTwoDenseCkptFourBest);

\draw[draw=black,fill=color_red_3_1,postaction={pattern=flexible hatch, pattern color=black}] (axis cs:0.04,0) rectangle (axis cs:0.12,\evalRecoveryCriteoSizeOgCkptFourAverage);
\draw[draw=black,fill=color_red_3_1,postaction={pattern=flexible hatch, pattern color=black}] (axis cs:1.04,0) rectangle (axis cs:1.12,\evalRecoveryCriteoSizeTwoSparseCkptFourAverage);
\draw[draw=black,fill=color_red_3_1,postaction={pattern=flexible hatch, pattern color=black}] (axis cs:2.04,0) rectangle (axis cs:2.12,\evalRecoveryCriteoSizeTwoSparseTwoDenseCkptFourAverage);

\draw[draw=black,fill=color_red_3_2,postaction={pattern=flexible hatch, pattern color=black}] (axis cs:0.12,0) rectangle (axis cs:0.2,\evalRecoveryCriteoSizeOgCkptFourWorst);
\draw[draw=black,fill=color_red_3_2,postaction={pattern=flexible hatch, pattern color=black}] (axis cs:1.12,0) rectangle (axis cs:1.2,\evalRecoveryCriteoSizeTwoSparseCkptFourWorst);
\draw[draw=black,fill=color_red_3_2,postaction={pattern=flexible hatch, pattern color=black}] (axis cs:2.12,0) rectangle (axis cs:2.2,\evalRecoveryCriteoSizeTwoSparseTwoDenseCkptFourWorst);

\draw[draw=black,fill=color_green_3_0,postaction={pattern=north west lines, pattern color=black}] (axis cs:0.2,0) rectangle (axis cs:0.28,\evalRecoveryCriteoSizeOgCkptEightBest);
\draw[draw=black,fill=color_green_3_0,postaction={pattern=north west lines, pattern color=black}] (axis cs:1.2,0) rectangle (axis cs:1.28,\evalRecoveryCriteoSizeTwoSparseCkptEightBest);
\draw[draw=black,fill=color_green_3_0,postaction={pattern=north west lines, pattern color=black}] (axis cs:2.2,0) rectangle (axis cs:2.28,\evalRecoveryCriteoSizeTwoSparseTwoDenseCkptEightBest);

\draw[draw=black,fill=color_green_3_1,postaction={pattern=north west lines, pattern color=black}] (axis cs:0.28,0) rectangle (axis cs:0.36,\evalRecoveryCriteoSizeOgCkptEightAverage);
\draw[draw=black,fill=color_green_3_1,postaction={pattern=north west lines, pattern color=black}] (axis cs:1.28,0) rectangle (axis cs:1.36,\evalRecoveryCriteoSizeTwoSparseCkptEightAverage);
\draw[draw=black,fill=color_green_3_1,postaction={pattern=north west lines, pattern color=black}] (axis cs:2.28,0) rectangle (axis cs:2.36,\evalRecoveryCriteoSizeTwoSparseTwoDenseCkptEightAverage);

\draw[draw=black,fill=color_green_3_2,postaction={pattern=north west lines, pattern color=black}] (axis cs:0.36,0) rectangle (axis cs:0.44,\evalRecoveryCriteoSizeOgCkptEightWorst);
\draw[draw=black,fill=color_green_3_2,postaction={pattern=north west lines, pattern color=black}] (axis cs:1.36,0) rectangle (axis cs:1.44,\evalRecoveryCriteoSizeTwoSparseCkptEightWorst);
\draw[draw=black,fill=color_green_3_2,postaction={pattern=north west lines, pattern color=black}] (axis cs:2.36,0) rectangle (axis cs:2.44,\evalRecoveryCriteoSizeTwoSparseTwoDenseCkptEightWorst);

\draw[draw=black,fill=color_blue_2_0] (axis cs:0.44,0) rectangle (axis cs:0.52,\evalRecoveryCriteoSizeOgCodedServerFour);
\draw[draw=black,fill=color_blue_2_0] (axis cs:1.44,0) rectangle (axis cs:1.52,\evalRecoveryCriteoSizeTwoSparseCodedServerFour);
\draw[draw=black,fill=color_blue_2_0] (axis cs:2.44,0) rectangle (axis cs:2.52,\evalRecoveryCriteoSizeTwoSparseTwoDenseCodedServerFour);

\draw[draw=black,fill=color_blue_2_1] (axis cs:0.52,0) rectangle (axis cs:0.6,\evalRecoveryCriteoSizeOgCodedServerTwo);
\draw[draw=black,fill=color_blue_2_1] (axis cs:1.52,0) rectangle (axis cs:1.6,\evalRecoveryCriteoSizeTwoSparseCodedServerTwo);
\draw[draw=black,fill=color_blue_2_1] (axis cs:2.52,0) rectangle (axis cs:2.6,\evalRecoveryCriteoSizeTwoSparseTwoDenseCodedServerTwo);
\end{axis}

\end{tikzpicture}
    \vspace{-0.3in}
    \caption{Time to fully recover a failed \server.}
    \label{fig:eval:recovery_time}
\end{figure}
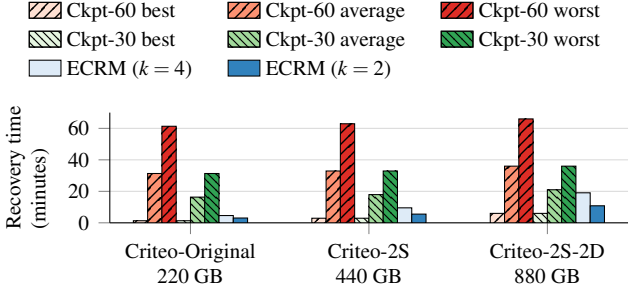

  \begin{figure}[t]
        \centering
        \newcommand{\figureheight}{1.4in}
        \newcommand{\figurewidth}{\linewidth}
        \begin{tikzpicture}[]
\tikzset{
        hatch distance/.store in=\hatchdistance,
        hatch distance=5pt,
        hatch thickness/.store in=\hatchthickness,
        hatch thickness=0.5pt
    }
\pgfplotsset{label style={font=\footnotesize}, 
             tick label style={font=\footnotesize},
             legend style={font=\footnotesize},
             every non boxed x axis/.append style={x axis line style=-},
             every non boxed y axis/.append style={y axis line style=-},
             axis lines=left,
             /pgfplots/ybar legend/.style={
                /pgfplots/legend image code/.code={%
                    \draw[##1,/tikz/.cd,yshift=-0.25em]
                    (0cm,0cm) rectangle (12pt,6pt);
                },
            },
        }

\begin{axis}[
height=\figureheight,
legend cell align={left},
legend columns=2,
legend style={at={(0.5,1.5)}, anchor=north, draw=none, /tikz/every even column/.append style={column sep=0.4cm}},
tick align=outside,
tick label style={/pgf/number format/assume math mode},
tick pos=left,
width=\figurewidth,
x grid style={white!69.01960784313725!black},
xmin=-0.205, xmax=2.655,
xtick={0.225,1.225,2.225},%
xticklabel style={align=center},
xticklabels={220,440,880},
xlabel={\dlrm size (GB)},
y grid style={gray, opacity=0.3},
ylabel style={at={(axis description cs:0.04,.5)},align=left},
ymajorgrids,
ylabel={Increase in\\training time (\%)},
ymin=0, ymax=35
]

\addlegendimage{ybar,ybar legend,draw=black,fill=color_red_2_0,postaction={pattern=flexible hatch, pattern color=black}};
\addlegendentry{\ckptSixty}
\addlegendimage{ybar,ybar legend,draw=black,fill=color_blue_2_0};
\addlegendentry{\sysKFour}
\addlegendimage{ybar,ybar legend,draw=black,fill=color_red_2_1,postaction={pattern=flexible hatch, pattern color=black}};
\addlegendentry{\ckptThirty}
\addlegendimage{ybar,ybar legend,draw=black,fill=color_blue_2_1};
\addlegendentry{\sysKTwo}

\draw[draw=black,fill=color_red_2_0,postaction={pattern=flexible hatch, pattern color=black}] (axis cs:-0.075,0) rectangle (axis cs:0.075,\evalNormalCriteoSizeOgCkptFour);
\draw[draw=black,fill=color_red_2_0,postaction={pattern=flexible hatch, pattern color=black}] (axis cs:0.925,0) rectangle (axis cs:1.075,\evalNormalCriteoSizeTwoSparseCkptFour);
\draw[draw=black,fill=color_red_2_0,postaction={pattern=flexible hatch, pattern color=black}] (axis cs:1.925,0) rectangle (axis cs:2.075,\evalNormalCriteoSizeTwoSparseTwoDenseCkptFour);

\draw[draw=black,fill=color_red_2_1,postaction={pattern=flexible hatch, pattern color=black}] (axis cs:0.075,0) rectangle (axis cs:0.225,\evalNormalCriteoSizeOgCkptEight);
\draw[draw=black,fill=color_red_2_1,postaction={pattern=flexible hatch, pattern color=black}] (axis cs:1.075,0) rectangle (axis cs:1.225,\evalNormalCriteoSizeTwoSparseCkptEight);
\draw[draw=black,fill=color_red_2_1,postaction={pattern=flexible hatch, pattern color=black}] (axis cs:2.075,0) rectangle (axis cs:2.225,\evalNormalCriteoSizeTwoSparseTwoDenseCkptEight);

\draw[draw=black,fill=color_blue_2_0] (axis cs:0.225,0) rectangle (axis cs:0.375,\evalNormalCriteoSizeOgCodedServerFour);
\draw[draw=black,fill=color_blue_2_0] (axis cs:1.225,0) rectangle (axis cs:1.375,\evalNormalCriteoSizeTwoSparseCodedServerFour);
\draw[draw=black,fill=color_blue_2_0] (axis cs:2.225,0) rectangle (axis cs:2.375,\evalNormalCriteoSizeTwoSparseTwoDenseCodedServerFour);

\draw[draw=black,fill=color_blue_2_1] (axis cs:0.375,0) rectangle (axis cs:0.525,\evalNormalCriteoSizeOgCodedServerTwo);
\draw[draw=black,fill=color_blue_2_1] (axis cs:1.375,0) rectangle (axis cs:1.525,\evalNormalCriteoSizeTwoSparseCodedServerTwo);
\draw[draw=black,fill=color_blue_2_1] (axis cs:2.375,0) rectangle (axis cs:2.525,\evalNormalCriteoSizeTwoSparseTwoDenseCodedServerTwo);
\end{axis}

\end{tikzpicture}
        \vspace{-0.2in}
        \caption{Training-time overhead. %
        }
        \label{fig:eval:normal_time}
  \end{figure}
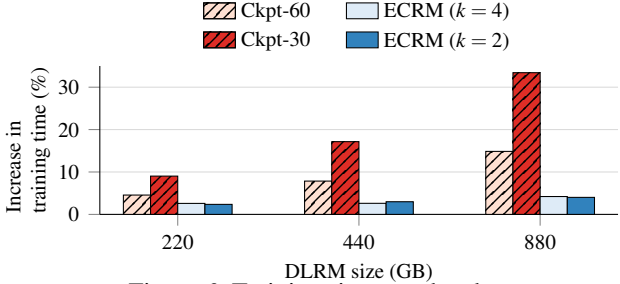
  \begin{figure}[t]
        \centering
        \newcommand{\figureheight}{1.4in}
        \newcommand{\figurewidth}{\linewidth}
        \begin{tikzpicture}[spy using outlines=
	{rectangle, magnification=3, width=2.2cm, height=1.0cm, connect spies, rounded corners, every spy on node/.append style={thick}}]

\pgfplotsset{label style={font=\small}, 
             tick label style={font=\small},
             legend style={font=\footnotesize},
             every non boxed x axis/.append style={x axis line style=-},
             every non boxed y axis/.append style={y axis line style=-},
             axis lines=left}

\definecolor{color0}{rgb}{0.12156862745098,0.466666666666667,0.705882352941177}
\definecolor{color1}{rgb}{1,0.498039215686275,0.0549019607843137}
\definecolor{color2}{rgb}{0.172549019607843,0.627450980392157,0.172549019607843}
\definecolor{color3}{rgb}{0.83921568627451,0.152941176470588,0.156862745098039}
\definecolor{color4}{rgb}{0.580392156862745,0.403921568627451,0.741176470588235}
\definecolor{color5}{rgb}{0.549019607843137,0.337254901960784,0.294117647058824}

\begin{axis}[
height=\figureheight,
legend cell align={left},
legend columns=2,
legend style={at={(0.5,1.2)}, anchor=south, draw=none,
            /tikz/every even column/.append style={column sep=0.2cm}},
tick align=outside,
tick label style={/pgf/number format/assume math mode},
tick pos=left,
width=\figurewidth,
x grid style={gray, opacity=0.3},
xmajorgrids,
xmin=10, xmax=240,%
xlabel={Time (minutes)},
y grid style={gray, opacity=0.3},
ymajorgrids,
ylabel style={align=left,at={(axis description cs:0.05,.5)}},
ylabel={Throughput\\(samples/sec)},
ymin=0, ymax=500000
]
        \input{figures/evaluation/data/normal/criteo-4x/tput}
        \coordinate (spypoint) at (axis cs:175,393000);
        \coordinate (magnifyglass) at (axis cs:175,160000);

\end{axis}
\spy [black,thick] on (spypoint)
  in node[fill=white] at (magnifyglass);

\end{tikzpicture}
        \vspace{-0.1in}
        \caption{Throughput of training \progressModel %
        }
        \label{fig:eval:normal_tput}
  \end{figure}
  \begin{figure}[t]
        \centering
        \newcommand{\figureheight}{1.4in}
        \newcommand{\figurewidth}{\linewidth}
        \begin{tikzpicture}[spy using outlines=
	{rectangle, magnification=2, width=1cm, height=1cm, connect spies, rounded corners}]

\pgfplotsset{label style={font=\small}, 
             tick label style={font=\small},
             legend style={font=\footnotesize},
             every non boxed x axis/.append style={x axis line style=-},
             every non boxed y axis/.append style={y axis line style=-},
             axis lines=left}

\definecolor{color0}{rgb}{0.12156862745098,0.466666666666667,0.705882352941177}
\definecolor{color1}{rgb}{1,0.498039215686275,0.0549019607843137}
\definecolor{color2}{rgb}{0.172549019607843,0.627450980392157,0.172549019607843}
\definecolor{color3}{rgb}{0.83921568627451,0.152941176470588,0.156862745098039}
\definecolor{color4}{rgb}{0.580392156862745,0.403921568627451,0.741176470588235}
\definecolor{color5}{rgb}{0.549019607843137,0.337254901960784,0.294117647058824}

\begin{axis}[
height=\figureheight,
legend cell align={left},
legend columns=2,
legend style={at={(0.5,1.2)}, anchor=south, draw=none,
            /tikz/every even column/.append style={column sep=0.2cm}},
tick align=outside,
tick label style={/pgf/number format/assume math mode},
tick pos=left,
width=\figurewidth,
x grid style={gray, opacity=0.3},
xmajorgrids,
xmin=10, xmax=240,%
xlabel={Time (minutes)},
y grid style={gray, opacity=0.3},
ymajorgrids,
ylabel style={align=left,at={(axis description cs:0.06,.5)}},
ylabel={Sample Number},
ymin=0, ymax=6586970112
]
        \input{figures/evaluation/data/normal/criteo-4x/progress}
\end{axis}
\end{tikzpicture}
        \vspace{-0.1in}
        \caption{Progress of training \progressModel %
        }
        \label{fig:eval:normal_progress}
  \end{figure}

\subsection{Performance during recovery} \label{sec:evaluation:results:recovery}
We first evaluate the performance of \sys and checkpointing in recovering from failure. As recovery time for checkpointing depends on when failure occurs (see \Section\ref{sec:background:checkpointing:overhead}), we show the best-, average-, and worst-case for checkpointing. The recovery performance of \sys and checkpointing is best compared in \Figure~\ref{fig:eval:recovery_tput}, which plots the throughput and training progress of \sys and \ckptThirty on \criteoTwoSparseTwoDense after a single \server failure at time 10 minutes. As the recovery performance of \ckptSixty is even worse than \ckptThirty, we omit it from the plots for clarity. \sys fully recovers faster than the average case for \ckptThirty, and, critically, maintains throughput within 6\%--12\% of that during normal operation during recovery. In contrast, \ckptThirty cannot perform new training iterations during recovery. As shown in the bottom plot of \Figure~\ref{fig:eval:recovery_tput}, which plots the time taken to reach a number of training samples, \sys's high throughput during recovery enables it to progress in training faster than even the best case for \ckptThirty.

\Figure\ref{fig:eval:recovery_time} shows the time it takes for \sys, \ckptThirty, and \ckptSixty to recover a failed \server. \sys with $k=4$ recovers 1.9--6.8$\times$ faster and 1.1--3.5$\times$ faster than the average case for \ckptSixty and \ckptThirty, respectively (and {up to 10.3$\times$} faster with $k=2$).  More importantly, unlike checkpointing, \sys enables training to continue during recovery with high throughput.

\paragraphOrTextbf{Effect of parameter $\mathbf{k}$ and \dlrm size.} \Figure~\ref{fig:eval:recovery_time} illustrates the discussion from \Section\ref{sec:discussion:k} that it takes longer for \sys to fully recover with higher value of parameter $k$. However, \sys maintains high throughput during recovery for each value of  $k$ (see \Figure~\ref{fig:eval:recovery_tput}).

\Figure\ref{fig:eval:recovery_time} also shows that the time to fully recover increases with \dlrm size for both \sys and checkpointing, as expected (see \Section\ref{sec:background:checkpointing:overhead} and \Section\ref{sec:discussion:k}). \sys's recovery time increases more quickly with \dlrm size than checkpointing due to the $k$-fold increase in data read and compute performed by a single \server in \sys when decoding. However, this does not significantly affect training in \sys because \sys can continue training during recovery with high throughput.

\paragraphOrTextbf{Effect of lock granularity.} Section~\ref{sec:technique:recovery:nonblocking} described \sys's approach of using granular locks to prevent race conditions during recovery. The number of \embeddingTableEntries covered by each lock (\ie the lock granularity) leads to a tradeoff between memory overhead in recovery for buffering updates and time overhead for switching locks.

To evaluate this tradeoff, we compare the full recovery time of \sys with $k=4$ when using a single lock throughout recovery, and that when using 10 locks during recovery. %
Using 10 locks increases recovery time by 23.3\%, 10\%, and 9.4\% for \criteoOg, \criteoTwoSparse, and \criteoTwoSparseTwoDense, respectively. It is important to note that, even when employing locks with finer granularity, and thus having longer overall recovery time, \sys can continue to provide high training throughput during recovery, whereas checkpointing requires training to pause during recovery. Switching locks during recovery in \sys involves (1) momentarily pausing training to synchronize \workers and \servers, and (2) copying updated \embeddingTableEntries from buffers to the original \embeddingTableEntries. The time overhead incurred from synchronization is constant regardless of \dlrm size, whereas buffer copying overhead will grow with \dlrm size. Thus, with larger \dlrm sizes, synchronization overhead is better amortized, reducing the overall overhead of lock switching for larger \dlrms.

\subsection{Performance during normal operation} \label{sec:evaluation:results:normal}
We now compare the performance of \sys and checkpointing during normal operation.

\Figure\ref{fig:eval:normal_time} shows the training-time overhead of \sys and checkpointing as compared to a system with no fault tolerance (and thus no overhead) in a four-hour training run. \sys reduces training-time overhead during normal operation by 71.3\%--88\% and 41.3\%--71.6\% compared to \ckptThirty and \ckptSixty, respectively. While the training-time overhead of checkpointing decreases with decreased checkpointing frequency, \Section\ref{sec:evaluation:results:recovery} showed that this came the expense of recovery performance. It is important to note that \sys's benefit over checkpointing grows with \dlrm size. For example, on the \criteoTwoSparseTwoDenseSizeGB \criteoTwoSparseTwoDense, \ckptThirty has training-time overhead of 33.4\%, while \sys has overheads of 4.2\% and 4\% with $k$ of 4 and 2, respectively. This illustrates the promise of \sys for future \dlrms, which are expected to grow in size.

\Figure\ref{fig:eval:normal_tput} plots the throughput of \sys and \ckptThirty compared to training with no fault tolerance (\noFT) on \progressModel. As shown in the inset, \sys has slightly lower throughput compared to \noFT, while \ckptThirty causes throughput to fluctuate from that equal to \noFT, to zero when writing a checkpoint. The effects of this fluctuation are shown in \Figure\ref{fig:eval:normal_progress}: \ckptThirty progresses slower than \sys.

\paragraphOrTextbf{Effect of parameter $\mathbf{k}$.} As described in \Section\ref{sec:discussion:k}, \sys has constant network bandwidth and CPU overhead during normal operation regardless of the value of parameter $k$. This is illustrated in Figures~\ref{fig:eval:normal_time}, \ref{fig:eval:normal_tput}, and \ref{fig:eval:normal_progress}, in which \sys has nearly equal performance with $k=2$ and $k=4$. 

We also measure the overhead of \sys with $k=10$ on a cluster twice the size as that in \Section\ref{sec:evaluation:setup} (to accommodate the higher value of $k$) and on a version of \criteoOg scaled up to have the same number of \entries per \server as in the original cluster. In this setting, \sys has training-time overhead of 0.5\%. This smaller overhead stems not from the increase in $k$, but from the decreased load on each \server due to the increased number of \servers. Nevertheless, this experiment shows that \sys can support high values of $k$.

\paragraphOrTextbf{Effect of \sys on load imbalance.}
{We next evaluate the effect of \sys's approach to parity placement on cluster load imbalance.} We measure load on each \server by counting the number of updates that occur on each \server when training \criteoOg.

When training without erasure coding, the most-heavily loaded \server performs 2.28$\times$ more \updates than the least-heavily loaded \server. In contrast, in \sys with $k=2$ and $k=4$, this difference in load is 1.64$\times$ and 1.58$\times$, respectively. This indicates that the increased load introduced by \sys leads to \textit{improved load balance.} Under \sys, parities corresponding to the \embeddingTableEntries of a given \server are distributed among all other \servers. Thus, the same amount of load that an individual \server experiences for non-parity \updates will also be distributed among the other \servers to update parities. While all \servers will experience increased load, the most-loaded \server in the absence of erasure coding is likely to experience the smallest increase in load due to the addition of erasure coding because all other \servers for whom it hosts parities have lower load. A similar argument holds for the least-loaded \server experiencing the largest increase in load. Hence, the expected difference in load between the most- and least-loaded \servers will decrease. Thus, while \sys doubles the total number of updates in the system, its impact is alleviated by improved load balancing provided by its approach to parity placement.

\paragraphOrTextbf{Effect of reduced \server resources.} As \sys introduces CPU and network bandwidth overhead on \servers during training, it is expected that \sys will have higher training-time overhead when \server CPU and network resources are limited.
We evaluate \sys in these settings by artificially limiting these resources when training \criteoOg.

To evaluate \sys with limited \server CPU resources, we replace the r5n.8xlarge \server instances described in \Section\ref{sec:evaluation:setup} with x1e.2xlarge instances, which have the same amount of memory, but $4\times$ less CPU cores. \sys's training-time overhead with $k=4$ is 11.1\% when using these instances, higher than that on the more-capable \servers (2.6\%).

To evaluate \sys with limited \server network bandwidth, we replace the r5n.8xlarge instances (which have 25 Gbps) described in \Section\ref{sec:evaluation:setup} with r5.8xlarge instances (which have 10 Gbps). \sys's training-time overhead with $k=4$ is 6.5\% on these bandwidth-limited instances, higher than that on the more-capable \servers (2.6\%).

Even on these resource-limited \servers, \sys still benefits from significantly improved performance during recovery compared to checkpointing and has training-time overhead comparable to \ckptThirty and slightly higher than \ckptSixty.

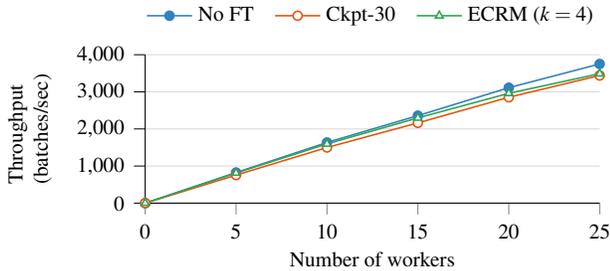
\begin{figure}[t]
    \newcommand{\figureheight}{1.4in}
    \newcommand{\figurewidth}{0.9\linewidth}
    \centering
    \begin{tikzpicture}
\pgfplotsset{label style={font=\footnotesize}, 
             tick label style={font=\footnotesize},
             legend style={font=\footnotesize},
             every non boxed x axis/.append style={x axis line style=-},
             every non boxed y axis/.append style={y axis line style=-},
             axis lines=left,
             /pgfplots/ybar legend/.style={
                /pgfplots/legend image code/.code={%
                    \draw[##1,/tikz/.cd,yshift=-0.25em]
                    (0cm,0cm) rectangle (12pt,6pt);
                },
            },
        }

\definecolor{color0}{rgb}{0.12156862745098,0.466666666666667,0.705882352941177}
\definecolor{color1}{rgb}{1,0.498039215686275,0.0549019607843137}
\definecolor{color2}{rgb}{0.172549019607843,0.627450980392157,0.172549019607843}
\definecolor{color3}{rgb}{0.83921568627451,0.152941176470588,0.156862745098039}

\begin{axis}[
height=\figureheight,
legend cell align={left},
legend columns=3,
legend style={at={(0.5,1.4)}, anchor=north, draw=none, /tikz/every even column/.append style={column sep=0.2cm}},
tick align=outside,
tick label style={/pgf/number format/assume math mode},
tick pos=left,
width=\figurewidth,
x grid style={white!69.01960784313725!black},
xlabel={Number of workers},
xmin=0, xmax=25,
xtick={0,5,10,15,20,25},
y grid style={gray, opacity=0.3},
ymajorgrids,
ylabel style={align=left,at={(axis description cs:-0.05,.5)}},
ylabel={Throughput\\(batches/sec)},
ymin=0, ymax=4000,
ytick style={color=black},
]
\addplot [semithick, color_blue_2_1, mark=*, mark size=1.7, mark options={solid}]
table {%
0	0
5	824
10	1635
15	2358
20	3111
25	3751
};
\addlegendentry{\noFT}
\addplot [semithick, color_orange_2_1, mark=*, mark size=1.7, mark options={solid,fill=white}]
table {%
0	0
5	755.7622284
10	1499.601024
15	2162.727348
20	2853.369287
25	3440.369076
};
\addlegendentry{\ckptThirty}
\addplot [semithick, color_green_2_1, mark=triangle*, mark size=1.7, mark options={solid,fill=white}]
table {%
0	0
5	813
10	1596
15	2297
20	2960
25	3489
};
\addlegendentry{\sysKFour}
\end{axis}

\end{tikzpicture}
    \vspace{-0.1in}
    \caption{Average training throughput with varying number of workers during normal operation}
    \label{fig:mem_line}
\end{figure}

\paragraphOrTextbf{Effect of number of \workers.} We additionally performed experiments with a different number of \workers in the system. We vary the number of \workers from 5 to 25 corresponding to 1$\times$ to 5$\times$ the number of \servers, and measure the average training throughput. Increasing the number of \workers in the system increases the rate at which gradients are generated, and thus the load on \servers in the system. 

\Figure\ref{fig:mem_line} shows the throughput of \sys, \ckptThirty, and training without any fault tolerance (\noFT) on \criteoOg. Compared to \noFT, \sys's overhead increases with the number of \workers from  1.4\% with 5 \workers to 2.6\% for 15 \workers, and finally to 7.5\% for 25 \workers. This trend is expected: as the number of \workers increases, gradients are sent to \servers at a higher rate, which increases the load on each \server. These more heavily-loaded \servers are more heavily affected by the increase in load due to updating parities in \sys than less-heavily loaded \servers. \Figure\ref{fig:mem_line} also shows the training throughput of \ckptThirty has a constant overhead of 9.0\%. \textit{The results show that even in settings with higher worker to server ratio, \sys maintains lower overhead than checkpointing during normal operation.}.

\paragraphOrTextbf{Benefit of \serverSideProp.} 
One motivation for \sys's approach of \serverSideProp is to efficiently update parities (see \Section\ref{sec:technique:server:server}). To illustrate this, we compare \sys to the naive alternative, \workerSideProp on \criteoOg. With $k = 4$, \workerSideProp has training-time overhead of 9.0\%, while \sys has only 2.6\% overhead, showing the benefit of \serverSideProp.
\subsection{Discussion} \label{sec:true_discussion}
\paragraphOrTextbf{Handling concurrent failures.} \label{sec:true_discussion:concurrent}
Recall from \Section\ref{sec:opportunity} that \sys leverages erasure codes parameterized with $r = 1$, that is, which can recover from a single server failure. This choice was informed by prior studies of cluster failures, which showed that single-node failures are the most common failure scenarios among groups of nodes across which data is encoded~\cite{rashmi2014hitchhiker}.

If higher fault tolerance is desired, \sys can be easily adapted to tolerate additional faults by using erasure codes with parameter $r > 1$. An alternative to this that still leverages $r = 1$ is to partition the overall cluster used in training into smaller groups of servers over which erasure coding with $r=1$ is performed, such that more than a single failure within each group is unlikely. Finally, \sys could also be adapted to take checkpoints at a much lower frequency than normal checkpointing schemes as a second layer of defense against concurrent failures. This final approach bears similarity to multi-level checkpointing~\cite{moody2010design}.

\paragraphOrTextbf{Synchronous training.} \label{sec:true_discussion:sync}
As described in \Section\ref{sec:background:dlrm}, many organizations deploying widely used recommendation systems use asynchronous training~\cite{jiang2019xdl,acun2020understanding}. We have built \sys atop XDL, an asynchronous training framework from Alibaba. However, \sys can also support synchronous training. Synchronous training adds a barrier after a certain number of training iterations in which \workers communicate gradients with one another and \servers, combine these gradients, and perform a single \update to each modified parameter. In such a synchronous framework, \sys would require that \parityEntries also be updated during this barrier so that they are kept consistent with training \updates. As this setting is not the focus of our work, we leave a full study and evaluation of \sys in synchronous settings to future work.

\paragraphOrTextbf{Checkpointing for model deployment.} \label{sec:true_discussion:checkpointing} In addition to checkpointing for fault tolerance, a system may also take checkpoints of a \dlrm for later deployment. These checkpoints are taken infrequently (\eg at the end of training), and thus add little overhead. \sys does not preclude the use of checkpointing for this purpose.
\section{Related Work} \label{sec:related}
\paragraphOrTextbf{\dlrm systems.} 
System support for recommendation model training and inference has recently received significant attention.
Solutions tailored toward understanding and improving \dlrm inference range from workload and system analysis~\cite{gupta2020architectural,lui2020understanding}, model-system codesign~\cite{eisenman2018bandana,gupta2020deeprecsys}, and specialized hardware support~\cite{jiang2020microrec,wilkening2021recssd}.

More recently, work has emerged for improving the performance of training \dlrms. For example, recent works from organizations that train large-scale \dlrms have described systems designed for training \dlrms~\cite{jiang2019xdl,naumov2019deep,zhao2019aibox,kalamkar2020optimizing,acun2020understanding}. Other works have focused on model-system codesign, such as reducing the sizes of \embeddingTables through compression and precision-reduction techniques~\cite{ginart2019mixed,huang2020mixed,yin2021tt}.

\sys differs from these works by its focus  on fault tolerance for \dlrm training and its novel use of erasure codes therein.

\paragraphOrTextbf{Checkpointing.} Computer systems have long used checkpointing for fault tolerance~\cite{koo1987checkpointing,daly2006higher,moody2010design}. Recent works optimize checkpointing in \nnShort training~\cite{nicolae2020deepfreeze,mohan2021checkfreq}, but do not focus on \dlrm training. In contrast, \sys leverages the unique characteristics of \dlrm training to use erasure codes for efficient fault tolerance. Other works have developed approximation-based checkpointing techniques to reduce the overhead of checkpointing in iterative machine learning training algorithms~\cite{qiao2019fault,chen2020efficient}. Unlike these approaches, \sys does not change the accuracy guarantees of the underlying training system.

Most closely related to \sys are the works of Maeng et al.~\cite{maeng2021cpr} and Eisenman et al.~\cite{eisenman2020check}, which propose techniques for reducing the overhead of checkpointing in \dlrm training. Maeng et al.~\cite{maeng2021cpr} propose to use partial recovery to reduce the overhead of rolling back a \dlrm during recovery; when a failure occurs, only the failed node rolls back to its most recent checkpoint. Eisenman et al.~\cite{eisenman2020check} propose a combination of incremental checkpointing and reducing the numerical precision of checkpointed parameters. Both of these works leverage techniques that can potentially reduce the accuracy of \dlrm training upon recovering from failure. While both works empirically demonstrate small accuracy drops, they cannot provide the same accuracy guarantees as the underlying \dlrm training system. \sys differs from these works in two regards: (1) \sys maintains the same accuracy guarantees as the underlying \dlrm training system on top of which it is built. This reduces uncertainty about whether the fault tolerance approach being used will deliver a model with high enough accuracy needed for deployment, and reduces data scientist effort needed in debugging model performance when there are multiple sources of potential inaccuracy present. (2) \sys leverages erasure coding to reduce the overhead of fault tolerance.

\paragraphOrTextbf{Erasure-coded systems.} Erasure codes have been widely used in storage systems, communication systems, and caching systems for various purposes, such as fault/loss tolerance, load balancing, and alleviation of slowdowns~\cite{patterson1988raid,rizzo1997effective,rashmi2016eccache,yan2017tiny}. To the best of our knowledge, \sys is the first approach of leveraging erasure codes for fault tolerance within \dlrm training systems. Of the more traditional uses of erasure codes described above, \sys bears greatest similarity erasure-coded storage systems. As described in \Section\ref{sec:opportunity}, some of the techniques employed by \sys (\eg rotating parity placement) are inspired by erasure-coded storage systems and are adapted to the unique properties of \dlrm training systems.

\textbf{Erasure coding in machine learning systems.} Recent work has applied coding-theoretic ideas to machine learning systems. These works primarily focus on alleviating the effects of transient slowdowns in \nn inference systems~\cite{kosaian2019parity} and in training certain classes of machine learning models (\eg~\cite{tandon2017gradient,lee2018speeding,dutta2018unified,yu2018lagrange}). In contrast, \sys imparts fault tolerance to \dlrm training, which differs significantly in model architecture and system design from the settings considered in these works.

\section{Conclusion} \label{sec:conclusion}
\sys is a new approach to fault-tolerant \dlrm training that employs erasure coding to overcome the downsides of checkpointing. \sys encodes the large embedding tables and \optimizer state in \dlrms, maintains up-to-date parities with low overhead, and enables training to continue during recovery, while maintaining the same accuracy guarantees as the underlying training system. Compared to checkpointing, \sys reduces training-time overhead in the absence of failures by up to 88\%, recovers from failures up to 10.3$\times$ faster, and allows training to proceed without pauses both during normal operation or recovery. 
While \sys's benefits come at the cost of additional memory requirements and load on the servers, the impact of these is alleviated by the fact that memory overhead is only fractional and that load gets evenly distributed. 
\sys shows the potential of erasure coding as a superior alternative to checkpointing for fault tolerance in training current and future \dlrms.


\begin{appendices}
\section{Cross-parameter consistency of \sys} \label{sec:appendix:consistency}
As described in \Section\ref{sec:discussion:consistency}, \sys guarantees that a recovered \dlrm represents one that could have been reached by asynchronous training, but does not guarantee that the recovered \dlrm represents a state that was truly experienced during recovery. We will next illustrate this by example and show how the guarantee above results in \sys providing the same consistency semantics as asynchronous training.

\newcommand{\suppEmbOne}{x\xspace}
\newcommand{\suppEmbTwo}{y\xspace}
\newcommand{\suppTime}{t\xspace}
\newcommand{\suppEmbOneT}{\suppEmbOne_{\suppTime}\xspace}
\newcommand{\suppEmbTwoT}{\suppEmbTwo_{\suppTime}\xspace}
\newcommand{\suppEmbOneTPlusOne}{\suppEmbOne_{\suppTime+1}\xspace}
\newcommand{\suppEmbTwoTPlusOne}{\suppEmbTwo_{\suppTime+1}\xspace}

Consider the timeline of events in Table~\ref{table:appendix:consistency} for training a \dlrm with \embeddingTableEntries $\suppEmbOne$ and $\suppEmbTwo$. We consider the state of the \dlrm to the combined state of each of these parameters.

\begin{table}[t]{
\centering
\caption{Order of events illustrating consistency of \sys atop asynchronous training.}
\label{table:appendix:consistency}
\begin{tabular}{p{0.07\linewidth}p{0.19\linewidth}p{0.18\linewidth}p{0.51\linewidth}}
\textbf{Time} & \textbf{Prev.\ State} & \textbf{New State} & \textbf{Event} \\ \hline & \\
0 & $\suppEmbOneT$, $\suppEmbTwoT$ & $\suppEmbOneTPlusOne$, $\suppEmbTwoT$ & \Entry $\suppEmbOne$ is updated from $\suppEmbOneT$ to $\suppEmbOneTPlusOne$ on Server \idxOne. \Entry and \optimizer difference is asynchronously propagated to the \server holding the parity for $\suppEmbOne$. \\ & & & \\
1                  & $\suppEmbOneTPlusOne$, $\suppEmbTwoT$ & $\suppEmbOneTPlusOne$, $\suppEmbTwoT$ & \Entry $\suppEmbTwoT$ is read from \Server \idxTwo. \\ & & &  \\
2  & $\suppEmbOneTPlusOne$, $\suppEmbTwoT$ & $\suppEmbOneTPlusOne$, $\suppEmbTwoTPlusOne$ & \Entry $\suppEmbTwo$ is updated from $\suppEmbTwoT$ to $\suppEmbTwoTPlusOne$ on Server \idxTwo. \Entry and \optimizer  difference is asynchronously propagated to the \server holding the parity for $\suppEmbTwo$. \\ & & & \\
3                   & $\suppEmbOneTPlusOne$, $\suppEmbTwoTPlusOne$ & $\suppEmbOneTPlusOne$, $\suppEmbTwoTPlusOne$ & The parity for entry $\suppEmbTwo$ is updated to reflect the change in $\suppEmbTwo$. \\ & & & \\
4                   & $\suppEmbOneTPlusOne$, $\suppEmbTwoTPlusOne$ & $\suppEmbOneTPlusOne$, $\suppEmbTwoTPlusOne$ & Server \idxOne fails, having not yet sent the difference for $\suppEmbOne$. \\ & & & \\
5                   & $\suppEmbOneTPlusOne$, $\suppEmbTwoTPlusOne$ & $\suppEmbOneT$, $\suppEmbTwoTPlusOne$ & Recovery process decodes $\suppEmbOne$.
\end{tabular}
}
\end{table}

As illustrated in the timeline in Table~\ref{table:appendix:consistency}, due to the asynchrony of \serverSideProp, the recovery process results in a \dlrm state \{$\suppEmbOneT$, $\suppEmbTwoTPlusOne$\} that was never experienced during training: in training, $\suppEmbOne$ was in state $\suppTime + 1$ before $\suppEmbTwoT$ was even read.

Though the \dlrm state recovered by \sys in the timeline above was never truly experienced during training, it \textit{is a \dlrm state that could have just as easily been experienced during asynchronous training}. Under asynchronous training, it would be just as valid for the event at time 0 to have been performed after the event at time 2, which would have resulted in the \dlrm state being \{$\suppEmbOneT$, $\suppEmbTwoTPlusOne$\}. Thus, the state recovered by \sys is still valid from the lens of asynchronous training.
\end{appendices}

\end{document}